\documentclass[journal]{IEEEtai}
\usepackage{amsmath,amsfonts}
\usepackage{algorithmic}
\usepackage{algorithm}
\usepackage{array}
\usepackage[caption=false,font=normalsize,labelfont=sf,textfont=sf]{subfig}
\usepackage{textcomp}
\usepackage{stfloats}
\usepackage{url}
\usepackage{verbatim}
\usepackage{graphicx}
\usepackage{cite}
\graphicspath{{figs/}}

\usepackage{bm}

\usepackage{amsmath}
\usepackage{amssymb}

\usepackage{algorithmic}
\usepackage{bbding}

\usepackage{multicol}
\usepackage{multirow}
\usepackage{booktabs}

\usepackage{xcolor, array, colortbl}

\definecolor{deepred}{rgb}{0.698,0.133,0.133}
\definecolor{blue}{rgb}{0,0,1}

\definecolor{orange}{rgb}{1,0.38,0}
\definecolor{beige}{rgb}{0.639,0.58,0.502}

\definecolor{lightgray}{rgb}{.91,.91,.91}

\usepackage{tikz,xcolor,hyperref}
\definecolor{lime}{HTML}{A6CE39}
\DeclareRobustCommand{\orcidicon}{
    \begin{tikzpicture}
    \draw[lime, fill=lime] (0,0) 
    circle [radius=0.16] 
    node[white] {{\fontfamily{qag}\selectfont \tiny ID}};    \draw[white, fill=white] (-0.0625,0.095) 
    circle [radius=0.007];    \end{tikzpicture}
    \hspace{-2mm}}
\foreach \x in {A, ..., Z}{
    \expandafter\xdef\csname orcid\x\endcsname{\noexpand\href{https://orcid.org/\csname orcidauthor\x\endcsname}{\noexpand\orcidicon}}
    }

\begin{document}

\title{Type-Balanced Contextual Learning for Incremental Named Entity Recognition}

\author{Duzhen Zhang,
        Yahan Yu,
        Xiuyi Chen,
        Chenxing Li,
    and Dong Yu\orcidA{},~\IEEEmembership{Fellow,~IEEE}
        \thanks{Manuscript created March, 2025. Duzhen Zhang and Yahan Yu contributed equally to this work. Chenxing Li is the corresponding author. Duzhen Zhang is with the Mohamed bin Zayed University of Artificial Intelligence, Abu Dhabi, UAE (E-mail: duzhen.zhang@mbzuai.ac.ae). Yahan Yu is with the Kyoto University, Kyoto, Japan (E-mail: yahan@nlp.ist.i.kyoto-u.ac.jp). Xiuyi Chen is with the Institute of Automation, Chinese Academy of Sciences, Beijing, China (E-mail: chenxiuyi2017@gmail.com). Chenxing Li is with the Tencent, AI Lab, Beijing, China (E-mail: chenxingli@tencent.com). Dong Yu is with the Tencent, AI Lab, Bellevue, WA 98004 USA (E-mail: dyu@global.tencent.com).}
        \thanks{The code is publicly available at \url{https://github.com/BladeDancer957/TBCL}.}
        }

\markboth{Journal of \LaTeX\ Class Files,~Vol.~14, No.~8, August~2021}%
{Shell \MakeLowercase{\textit{et al.}}: A Sample Article Using IEEEtran.cls for IEEE Journals}

\maketitle

\begin{abstract}

Incremental Named Entity Recognition (INER) stands as a pivotal task in information extraction, emphasizing the successive identification of new entity types within unstructured text. 
Faced with the continuous influx of entity types, INER grapples with two significant challenges: the widespread issue of catastrophic forgetting and the unique shift issue of the non-entity type semantics. 
While pseudo-labeling-based INER methods have proven effective in addressing these challenges, a previously overlooked issue arises: \textit{the biased context problem}. 
Our analysis shows that, in new sentences, the contextual associations of tokens representing old entity types exhibit a significantly stronger bias towards new entity types compared to their contexts in old sentences. 
This tendency intensifies the degradation of old knowledge while promoting the overfitting of new knowledge. 
To solve this biased context, we propose a Type-Balanced Contextual Learning (TBCL) method, featuring a sentence-duplet learning scheme and a contextual consistency loss. 
This approach offers a fresh perspective for INER through context analysis. 
Extensive experiments across ten INER settings on three highly recognized datasets showcase the efficacy of our TBCL method, highlighting its proficiency in resolving the biased context issue inherent in pseudo-labeling based INER approaches.
\end{abstract}

\begin{IEEEImpStatement}
Incremental Named Entity Recognition (INER) allows AI systems to continuously identify emerging entities in text, crucial for virtual assistants, automated healthcare records, and dynamic information platforms. 
Traditional methods often forget prior knowledge and overfit new data due to biased context. 
Our Type-Balanced Contextual Learning (TBCL) method corrects this by balancing old and new contextual information, ensuring stable learning over time. 
Extensive evaluations across benchmark datasets show TBCL consistently surpasses prior state-of-the-art approaches, offering more accurate, reliable, and adaptable AI models. This advancement improves AI's practical utility in healthcare, business intelligence, and education (social impact), reduces costly errors and inefficiencies (economic impact), and sets a new standard for sustainable incremental learning in real-world AI applications.
\end{IEEEImpStatement}


\section{Introduction}

\IEEEPARstart{N}{amed} Entity Recognition (NER) is crucial for information extraction, supporting various downstream tasks such as question answering~\cite{DBLP:conf/acl/LiYSLYCZL19,DBLP:conf/emnlp/LongprePCRD021} and web search queries~\cite{DBLP:conf/sigir/FetahuFRM21,DBLP:conf/sigir/GuoXCL09,DBLP:conf/aaai/MokhtariMY019,DBLP:conf/kdd/ZhangJD0YCTHWHC21}. NER focuses on identifying entities from unstructured text and classifying them into predefined categories (\emph{e.g.,} Person, Location) or as the non-entity type.
In the traditional fully-supervised NER framework~\cite{neural_for_nlp}, all predefined entity types are fixed and learned simultaneously during training, with the assumption that no new types will appear at the testing stage. Nevertheless, real-world scenarios frequently demand the discovery of novel entity types dynamically. For instance, virtual voice assistants like Alexa need to recognize new types (\emph{e.g.,} Genre, Actor) to understand emerging intents (\emph{e.g.,} GetMovie)~\cite{monaikul2021continual}. Conventional NER paradigms struggle when confronted with new entity types not encountered at the training stage. A simple strategy involves fine-tuning the NER model exclusively with new entity types samples, referred to as Incremental Named Entity Recognition (INER)\cite{continual_lifelong,lifelong_learning,monaikul2021continual}.
Effective INER methods~\cite{monaikul2021continual,xia2022learn,zheng2022distilling} are needed to incrementally update the model using only training samples of new entity types. However, INER faces two primary issues: catastrophic forgetting~\cite{catastrophic_1,catastrophic_2,catastrophic_3,catastrophic_4} and the semantic shift of the non-entity type~\cite{zhang2023task}.

\begin{figure}[t]
\centering
  \includegraphics[width=1.0\linewidth]{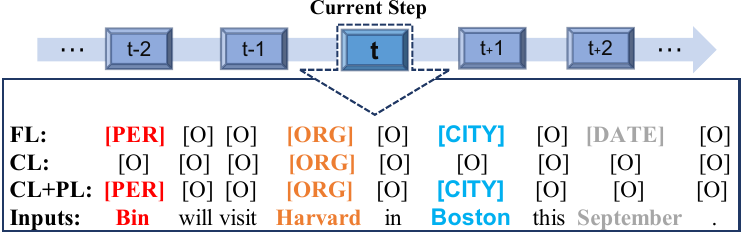} 
    \caption{A basic example of INER scenario, where \textbf{FL}, \textbf{CL}, and \textbf{PL} represent Full ground-truth Labels, Current ground-truth Labels, and Pseudo Labels, respectively. Previous entity types (\emph{e.g.,} \textcolor{red}{\textbf{[PER]}} (\textbf{Person}), \textcolor{blue}{\textbf{[CITY]}} (\textbf{City})) and future entity type (\emph{e.g.,} \textcolor{gray}{\textbf{[DATE]}} (\textbf{Date})) are marked as the non-entity type (\emph{i.e.}, \textcolor{black}{[O]}) in the current step $t$ where \textcolor{orange}{\textbf{[ORG]}} (\textbf{Organization}) is the current entity type being learned, leading to the shift of the non-entity type semantics (the second row \textbf{CL}). Using PL produced by the previous model to augment CL as the current classification target can effectively address the issue of semantic shift (the third row \textbf{CL+PL}).
}
\label{fig:motivation}
\end{figure}

The first challenge, catastrophic forgetting \cite{catastrophic_1,catastrophic_2,catastrophic_3,catastrophic_4}, is a widespread issue in incremental learning methodologies. This problem arises when model weights are modified to align with the optimal parameter space for newly added entity types, while lacking access to those encountered earlier. As a result, the NER model often forgets previously acquired knowledge of old entity types as it acquires information about new ones.
The second challenge, the semantic shift of the non-entity type \cite{zhang2023task}, is specific to INER. In contrast to conventional NER paradigms, where the non-entity type only includes tokens unrelated to any entity type, INER employs a distinct tagging approach. In INER, both previously learned and future entity types are tagged as the non-entity type at the current step, resulting in what we term the semantic shift of the non-entity type. As shown in Figure~\ref{fig:motivation} (the second row, \textbf{CL}), old types \textbf{Person} and \textbf{City} (learned in previous steps like $t-1$, $t-2$, $t-3$, etc.) and the future type \textbf{Date} (to be learned in future steps like $t+1$, $t+2$, $t+3$, etc.) are all masked as the non-entity type in the current step $t$, where \textbf{Organization} is the type currently being learned. If measures are not implemented to distinguish tokens associated with old entity types from the actual non-entity type, this phenomenon of semantic shift might worsen catastrophic forgetting.

\begin{figure}[t]
\centering
  \includegraphics[width=1.0\linewidth]{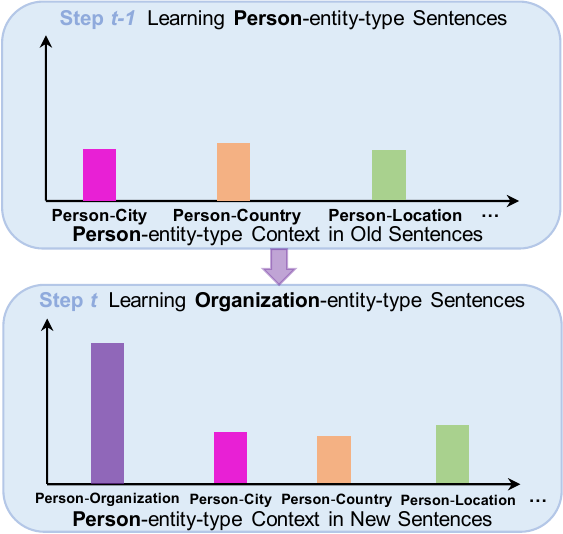} 
    \caption{Depiction of biased contextual relationships between the old-entity-type and new-entity-type tokens in the new sentences. At step $t-1$, when the NER model learns \textbf{Person}-entity-type sentences, the context of \textbf{Person} includes various entity types (\emph{e.g.}, \textbf{Person}-\textbf{City}, \textbf{Person}-\textbf{Country}, \textbf{Person}-\textbf{Location}). However, at step $t$, when the model learns the new \textbf{Organization}-entity-type sentences, the context of \textbf{Person} predominantly comprises \textbf{Person}-\textbf{Organization} pairs. Consequently, a new-entity-type-biased context for the old-entity-type (\textbf{Person}) tokens emerges in the new-entity-type (\textbf{Organization}) sentences, exacerbating the old-entity-type forgetting and new-entity-type overfitting issues.
}
\label{fig:context}
\end{figure}

Early INER methods \cite{monaikul2021continual,xia2022learn,zheng2022distilling} typically employ Knowledge Distillation (KD) \cite{hinton2015distilling} to address the problem of catastrophic forgetting. These approaches transfer output probability distribution from the previous model to the current model, thereby retaining the previously learned knowledge of old entity types. However, they overlook the shift issue of the non-entity type semantics specific to INER. To overcome this limitation, RDP \cite{zhang2023task}, a pseudo-labeling based INER method, is proposed. 
This method tackles the semantic shift issue by augmenting the current ground-truth labels with pseudo labels, resulting in more informative labels as the classification target for the current step (the third row \textbf{CL+PL} in Figure \ref{fig:motivation}). 
These pseudo labels are derived from predictions of the old model. 
By integrating KD and pseudo-labeling, RDP \cite{zhang2023task} attains State-Of-The-Art (SOTA) performance in INER.

Although RDP \cite{zhang2023task} effectively addresses the two primary challenges of INER, it introduces another challenge. 
Since NER is considered a context-dependent token-level classification task, we conducted a contextual analysis of entity types after pseudo-labeling \cite{zhao2022rbc}. 
As shown in Figure \ref{fig:context}, in step $t-1$, when the model learns the \textbf{Person} entity type, we observed that the co-occurrence frequency of \textbf{Person} with each of the pseudo-labeled old entity types (\emph{e.g.}, \textbf{City}, \textbf{Country}, \textbf{Location}) in the old sentences is almost the same. 
However, by step $t$, as the model learns the \textbf{Organization} entity type, the co-occurrence frequency between \textbf{Person} and \textbf{Organization} is greatly higher than with other entity types in the new sentences. 
We term this challenge the \textbf{biased context problem}, indicating that the context of old-entity-type tokens in the novel sentences exhibits a stronger bias towards new entity types compared to that in the old sentences. 
This bias can lead to a pronounced exacerbation of old-entity-type forgetting and new-entity-type overfitting.

To rectify this context bias, we introduce a Type-Balanced Contextual Learning (TBCL) method, which integrates a sentence-duplet learning scheme with a contextual consistency loss. This method aims to mitigate the impact of biased context information from old-entity-type tokens in the new incremental sentences. The sentence-duplet learning scheme utilizes pairs of sentences: the original sentence containing the new-entity-type tokens and a modified version with those tokens removed. This approach helps correct the context of old entity types in relation to the new ones. Additionally, the contextual consistency loss enforces a restriction on the context of old entity types within each sentence pair, ensuring consistency across contexts.

We highlight our core contributions as follows:
\begin{itemize}
    \item We pioneer the exploration of biased context in the INER scenario and introduce the TBCL method to address it. This method aims to mitigate the forgetting of old entity types while preventing the overfitting of new ones.
    
    \item We devise a novel sentence-duplet learning scheme and a contextual consistency loss, modifying how old entity types are contextualized with respect to new ones.
    
    \item Comprehensive experiments conducted on ten INER settings across three highly recognized datasets confirm the efficacy of the proposed TBCL method. The outcomes reveal that our method surpasses several existing INER methods and achieves new SOTA performance in INER.
\end{itemize}

\section{Related Work}

INER, an emerging field, has seen limited exploration in recent literature, with few papers tackling this issue. We begin this section with a summary of the latest progress in incremental learning, followed by a detailed analysis of current methods applied to INER.

\subsection{Incremental Learning}

Incremental learning is a methodology designed to continually absorb new information across successive tasks, with strategies to avoid catastrophic forgetting of earlier learnings, as noted in recent literature \cite{lifelong_machine,dong2023federated}. The techniques used in incremental learning fall into three categories: regularization-based, dynamic architecture-based, and replay-based methods. 
Regularization-based methods restrict adjustments to model weights \cite{EWC, MAS, SI, ONLINE-EWC}, intermediate features \cite{CLASSIFIER}, or the probabilities of outputs \cite{LWF} to preserve past knowledge.
Dynamic architecture-based methods \cite{PACKNET, DEN, DA} adjust the model's architecture dynamically to allocate resources effectively across various tasks, thereby stabilizing the learning process as new tasks are introduced.
Replay-based methods \cite{icarl, GEM, DGR} merge previously acquired or synthetically generated samples into current training sessions, balancing the retention of old information with the acquisition of new insights.

\subsection{INER}

Traditional NER typically classifies each token in a sequence into a standard set of entity types (\emph{e.g.,} Person, Organization, Location) or as the non-entity type \cite{ma2016end}. The focus has largely been on the development of various neural network architectures to enhance NER capabilities in a seamless, end-to-end process \cite{li2020survey}. Examples include techniques based on BiLSTM-CRF \cite{DBLP:conf/naacl/LampleBSKD16} and BERT \cite{kenton2019bert}. Yet, in practical applications, NER systems often encounter emerging new entity types that necessitate continuous model updates without complete retraining \cite{zhang2023decomposing,ma2023learning,chen2023skd,zhang2023continual}. To address this challenge, INER merges the incremental learning approach with traditional NER methods.

INER, still in its nascent stages, has seen only a limited number of studies tackling its specific challenges \cite{monaikul2021continual,xia2022learn,zheng2022distilling,zhang2025exploring,yu2024flexible,zhang2023task}. 
Among them, ExtendNER~\cite{monaikul2021continual} leverages KD techniques, wherein a teacher model (\emph{i.e.}, the previous model) passes on output probabilities to a student model (\emph{i.e.}, the newly updated model). L\&R~\cite{xia2022learn} presents a dual-phase architecture known as Learn-and-Review. In the learning phase, it adopts a method similar to ExtendNER's KD. In the review phase, it utilizes a rehearsal-based strategy, enhancing the current training dataset by artificially generated previous entity type samples. Meanwhile, CFNER~\cite{zheng2022distilling} innovates a causal framework designed to distill causal effects specifically from the non-entity type, adding a new dimension to traditional methods. 
Despite the significant advancements made by these INER methods in reducing catastrophic forgetting, they still fall short in effectively handling the shift issue in the non-entity type semantics \cite{zhang2023task}.

To remedy this defect, a pseudo-labeling based INER method called RDP \cite{zhang2023task} is proposed. 
This method develops a prototypical pseudo-label approach for classification, using the distances between token embeddings and type-wise prototypes to adjust the output probabilities of the previous model. This refinement helps to decrease prediction errors from the previous model and effectively handles the semantic shift issue. Combining KD with pseudo-labeling, RDP method achieves SOTA performance in the INER domain. 

However, we identified a specific biased context issue with this method, which exacerbates the forgetting of old entity types and leads to overfitting of new ones. To this end, we introduce a TBCL method, incorporating a sentence-duplet learning scheme and a contextual consistency loss. This approach is specifically tailored for the pseudo-labeling based INER method, further enhancing its performance by providing a more balanced learning context that mitigates these biases.

\section{Method}

In this section, we begin by outlining the definition of the INER problem. Subsequently, we provide an overview of the pseudo-labeling based INER method: RDP \cite{zhang2023task}. Finally, we delve into the detailed introduction of our TBCL approach.

\subsection{Problem Formulation}

Building on earlier studies \cite{monaikul2021continual,xia2022learn,zheng2022distilling,zhang2023task}, INER focuses on sequentially training a model across multiple steps $t=1,2,...,T$, each associated with a specific dataset $\mathcal{D}^1, \mathcal{D}^2, ..., \mathcal{D}^T$. 
Training at each step involves the current model $\mathcal{M}^t$ using only its corresponding dataset $\mathcal{D}^t$, which leads to the potential \textbf{catastrophic forgetting} of previously learned entity types from datasets $\mathcal{D}^1, \mathcal{D}^2, ..., \mathcal{D}^{t-1}$. 
Specifically, dataset $\mathcal{D}^t$ contains multiple training samples $(X^t, Y^t)$, with $X^t$ representing a sequence of input tokens (\emph{i.e.}, sentence) and ${Y}^t$ the corresponding labels in one-hot format.
These labels pertain only to the current entity types $\mathcal{E}^t$ relevant to step $t$. 
Labels from prior entity types $\mathcal{E}^{1:t-1}$ or future entity types $\mathcal{E}^{t+1:T}$ are merged into the non-entity type $e_{o}$, which introduces the issue of \textbf{semantic shift}. 
It's essential to emphasize that entity types are mutually exclusive across steps; that is, $\mathcal{E}^i \cap \mathcal{E}^j = \emptyset$ for $i \neq j$. 
The term $E^t=\text{card}(\mathcal{E}^t)$ denotes the cardinality of the entity types at the current step. 
The standard structure of the current model $\mathcal{M}^t$ includes an encoder $F^t$ and a linear softmax classifier $C^t$. 
Given the current dataset $\mathcal{D}^t$ and the old model $\mathcal{M}^{t-1}$, the objective with INER is to train the current (new) model $\mathcal{M}^t$ to effectively recognize all entity types $\mathcal{E}^{1:t}$ encountered up to the current step. 
{Let $X^t=(x^t_1,\ldots,x^t_{|X^t|})$ denote a sentence at step $t$. 
We define
the label space at step $t$ as $\mathcal{C}^t=\{e_o\}\cup \mathcal{E}^{1:t}$, where
$e_o$ denotes the non-entity type and $|\mathcal{C}^t|=1+E^1+\cdots+E^t$.
The one-hot ground-truth label matrix is denoted by
$Y^t\in\{0,1\}^{|X^t|\times|\mathcal{C}^t|}$, and the prediction matrix
produced by the current model is denoted by
$\widehat{Y}^t=\mathcal{M}^t(X^t;\Theta^t)\in[0,1]^{|X^t|\times|\mathcal{C}^t|}$.
Here, $Y^t_{i,e}$ and $\widehat{Y}^t_{i,e}$ denote the ground-truth label and
predicted probability of the $i$-th token for class $e$, respectively.} The cross-entropy loss for the current step $t$ is calculated as follows:
\begin{equation}
\mathcal{L}_{ce}(X^t,Y^t;\Theta^t)
=
-\frac{1}{|X^t|}
\sum_{i=1}^{|X^t|}
\sum_{e\in\mathcal{C}^t}
Y^t_{i,e}\log \widehat{Y}^t_{i,e},
\label{cross_entropy_loss}
\end{equation}
{where $\Theta^t$ denotes the trainable parameters of the current model $\mathcal{M}^t$.}

\subsection{Pseudo-labeling Based INER}

To mitigate issues of catastrophic forgetting and semantic shift, a pseudo-labeling based INER method, RDP \cite{zhang2023task}, is proposed. This approach uses the KD technique to preserve prior knowledge and designs a prototypical pseudo-label strategy for creating high-quality pseudo labels.

\subsubsection{KD} 

The KD technique alleviates catastrophic forgetting by transferring predicted probabilities from the previous model $\mathcal{M}^{t-1}$ to the current model $\mathcal{M}^t$. The objective function for distilling this probability distribution is formulated as follows:
\begin{equation}
\mathcal{L}_{\text{kd}}(X^t,\widehat{Y}^{t-1};\Theta^t)
=
-\frac{1}{|X^t|}
\sum_{i=1}^{|X^t|}
\sum_{e\in\mathcal{C}^{t-1}}
\widehat{Y}^{t-1}_{i,e}
\log \widehat{Y}^{t}_{i,e},
\label{kl_loss}
\end{equation}
{where $\widehat{Y}^{t-1}=\mathcal{M}^{t-1}(X^t)\in[0,1]^{|X^t|\times|\mathcal{C}^{t-1}|}$
and $\widehat{Y}^{t}=\mathcal{M}^{t}(X^t;\Theta^t)\in[0,1]^{|X^t|\times|\mathcal{C}^{t}|}$.
Since the old model only predicts over $\mathcal{C}^{t-1}$, the KD loss is
computed only on the old label space $\mathcal{C}^{t-1}$. That is, only the
entries of $\widehat{Y}^{t}$ corresponding to $\mathcal{C}^{t-1}$ are used.}

\subsubsection{Prototypical Pseudo-label Strategy} 

To tackle the semantic shift issue of the non-entity type, RDP introduces a prototypical pseudo-label strategy \cite{zhang2023task}. The goal is to utilize the predictions from the previous model for non-entity type tokens, treating them as indicators of their true entity type, particularly when they align with any of the old entity types. 
{We denote the pseudo-label target matrix for the current step by
$\widetilde{Y}^t\in\{0,1\}^{|X^t|\times|\mathcal{C}^t|}$. This target is computed
from the one-hot ground-truth label matrix $Y^t$ and the old-model prediction
$\widehat{Y}^{t-1}=\mathcal{M}^{t-1}(X^t)$ as follows:}
\begin{equation}
\widetilde{Y}^t_{i,e} =
\begin{cases}
1, & \text{if } Y^t_{i,e_o}=0
\text{ and } e=\arg\max_{e'\in \mathcal{E}^t}Y^t_{i,e'},\\
1, & \text{if } Y^t_{i,e_o}=1
\text{ and } e=\arg\max_{e'\in\mathcal{C}^{t-1}}\widehat{Y}^{t-1}_{i,e'},\\
0, & \text{otherwise}.
\end{cases}
\label{eq:pseudo_bis}
\end{equation}
{In this formulation, $\widetilde{Y}^t_{i,e}=1$ indicates that the $i$-th token is
assigned to class $e$. 
If the token is labeled as a current entity type, the
target follows the ground-truth label (the first row of Equation~(\ref{eq:pseudo_bis})). 
Otherwise, if it is labeled as the
non-entity type $e_o$, the target is generated from the old-model prediction
over the old label space $\mathcal{C}^{t-1}$ (the second row).}

\begin{figure*}[t]
\centering
  \includegraphics[width=1.0\linewidth]{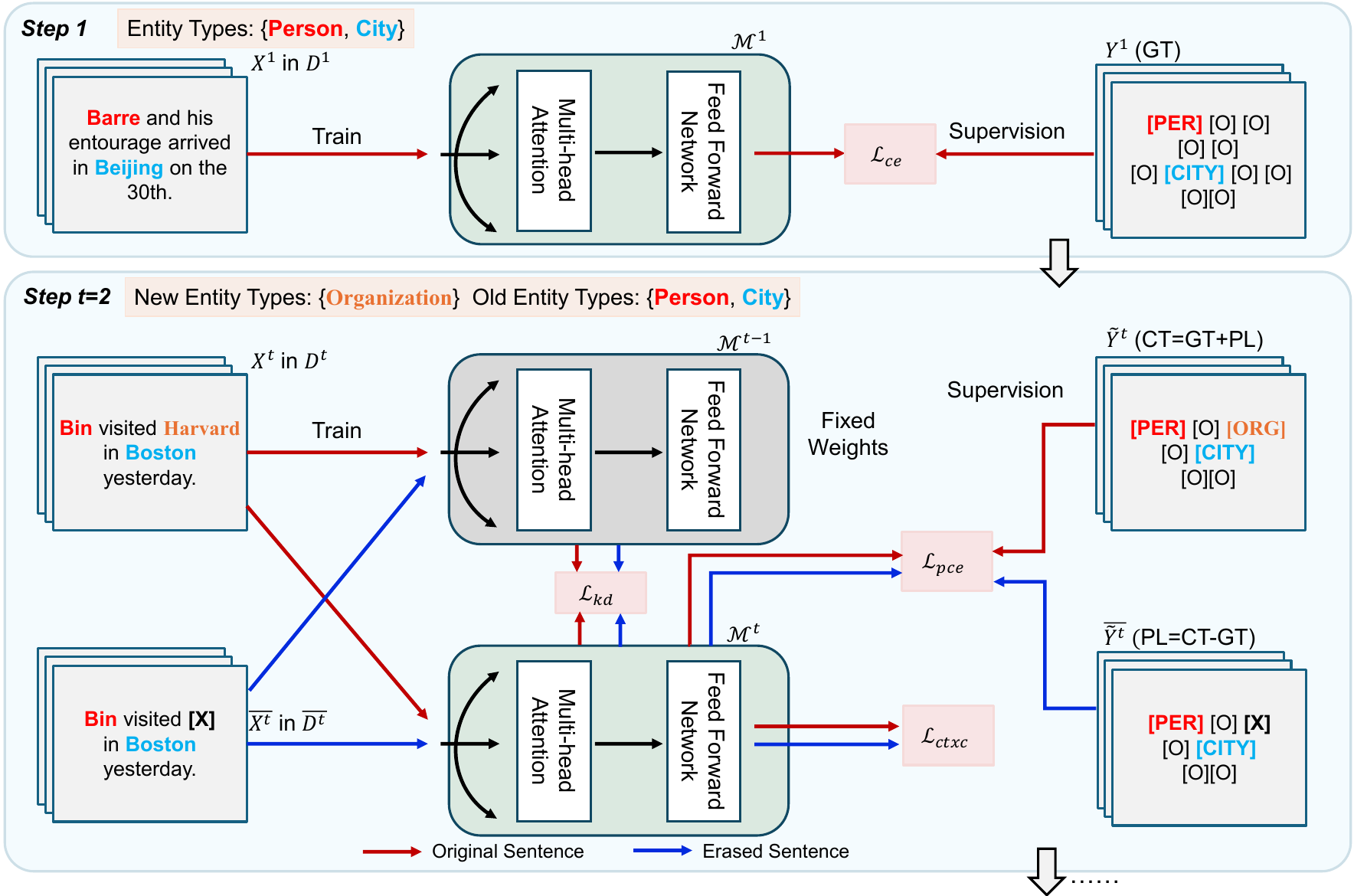} 
    \caption{Illustration of our TBCL method. Initially, at step $1$, the NER model is built from scratch with cross-entropy loss $\mathcal{L}_{\text{ce}}$ as the objective on dataset $\mathcal{D}^1$. In subsequent steps (\emph{e.g.}, at step $t$=$2$), we begin by obtaining sentence-duplets ($\mathcal{D}^t$, $\overline{\mathcal{D}^t}$) and update the model through our innovative sentence-duplet learning scheme with the pseudo-labeling cross-entropy loss $\mathcal{L}_{\text{pce}}$ and the KD loss $\mathcal{L}_{\text{kd}}$ and our unique contextual consistency loss $\mathcal{L}_{\text{ctxc}}$. Here, GT stands for current Ground-Truth labels, CT denotes Current Target, PL represents Pseudo Labels, [O] denotes the non-entity type, and [X] indicates an Erased Token.}
\label{fig:main}
\end{figure*}

To reduce prediction errors stemming from the old model and generate more accurate pseudo labels, RDP delve deeper by leveraging the distances between token embeddings and type-specific prototypes to adjust the output probabilities of the old model \cite{zhang2023task}. Equation (\ref{eq:pseudo_bis}) can be revised as follows:
\begin{equation}
\widetilde{Y}^t_{i,e} =
\begin{cases}
1, & \text{if } Y^t_{i,e_o}=0
\text{ and } e=\arg\max_{e'\in \mathcal{E}^t}Y^t_{i,e'},\\
1, & \text{if } Y^t_{i,e_o}=1
\text{ and } e=\arg\max_{e'\in\mathcal{C}^{t-1}}
\omega^t_{i,e'}\widehat{Y}^{t-1}_{i,e'},\\
0, & \text{otherwise}.
\end{cases}
\label{eq:pseudo_proto}
\end{equation}
{Here, $\omega^t_{i,e}$ represents the type-wise prototypical weight, quantifying the distance between the embedding of token $x^t_i$ and the type-specific prototype $\eta^e$.} 
This is expressed as follows:
\begin{equation}
\omega^t_{i,e} =
\frac{
\exp(-\|F^{t-1}(x^t_i)-\eta^e\|)
}{
\sum_{e'\in\mathcal{C}^{t-1}}
\exp(-\|F^{t-1}(x^t_i)-\eta^{e'}\|)
},
\label{prototypocal_weight}
\end{equation}
{where $F^{t-1}(x^t_i)$ denotes the embedding of token $x^t_i$ produced by the
encoder $F^{t-1}$ of the old model $\mathcal{M}^{t-1}$, and $\eta^e$ represents the
prototype for class $e$.} 
When the token embedding $F^{t-1}(x^t_i)$ is isolated from the prototype ${\eta}^{e}$ (\emph{i.e.}, the feature centroids of type $e$), it suggests that the learned feature might be considered an outlier. Consequently, this method reduces its likelihood of being classified into type $e$ and vice versa. To obtain the prototypes $\eta^e$, it begins by extracting coarse pseudo labels derived from the highest output probabilities predicted by the old model for all non-entity type tokens in the current dataset $\mathcal{D}^t$. Then, it computes the average embeddings of tokens identified as $e$ to establish the prototype $\eta^e$.

{The pseudo-labeling cross-entropy loss, computed utilizing these constructed pseudo labels, is formulated as follows:}
\begin{equation}
\mathcal{L}_{\text{pce}}(X^t,\widetilde{Y}^t;\Theta^t)
=
-\frac{1}{|X^t|}
\sum_{i=1}^{|X^t|}
\sum_{e\in\mathcal{C}^{t}}
\widetilde{Y}^t_{i,e}\log \widehat{Y}^t_{i,e}.
	\label{entropy_pseudo_label}
\end{equation}

{Overall, the objective function in the RDP method \cite{zhang2023task} is formulated as follows:}
\begin{equation}
\mathcal{L}_{\text{rdp}}(X,\widetilde{Y},\widehat{Y}^{old};\Theta^t)
=
\mathcal{L}_{\text{pce}}(X,\widetilde{Y};\Theta^t)
+
\alpha\mathcal{L}_{\text{kd}}(X,\widehat{Y}^{old};\Theta^t),
\label{gsc}
\end{equation}
{where $\widetilde{Y}$ is the pseudo-label target corresponding to input $X$, and
$\widehat{Y}^{old}=\mathcal{M}^{t-1}(X)$ is the old-model prediction on the same input.
The hyperparameter $\alpha$ balances the importance of the two loss terms.}

\subsection{TBCL}

While the aforementioned pseudo-labeling based INER method, RDP \cite{zhang2023task}, effectively mitigates catastrophic forgetting and semantic shift problems, the updated NER model it produces encounters issues with biased context.
For instance, at the $t$-th step (illustrated in Figure \ref{fig:context}), the newly added sentences predominantly consist of the new-entity-type-related context for the old-entity-type tokens, which exacerbates problems related to forgetting old entity types and overfitting to new ones.
To reduce the biased context correlation between old and new entity types in RDP \cite{zhang2023task} and to further improve its performance, we introduce a TBCL method. This method incorporates a sentence-duplet learning scheme and a contextual consistency loss, as depicted in Figure \ref{fig:main}.

\subsubsection{Sentence-duplet Learning Scheme}

Regarding the context related to new entity types, we notice that the contextual information of old entity type tokens within the new incremental sentences tends to be biased towards tokens of new entity types (as illustrated in Figure~\ref{fig:context}). To ensure the incremental learning of a NER model less susceptible to this intertwined new-entity-type context, we initially address the biased context among new and old entity types within these new sentences by removing the new-entity-type tokens from the original sentence (as illustrated in Figure~\ref{fig:main}). 
{At the $t$-th step, for each original sentence $X^t\in \mathcal{D}^t$, we construct its
erased version $\overline{X^t}$ by removing the new-entity-type tokens. The
pseudo-label targets for $X^t$ and $\overline{X^t}$ are denoted by $\widetilde{Y}^t$
and $\overline{\widetilde{\bm{Y}}^t})$, respectively. 
The corresponding predictions of the old
model are denoted by $\widehat{Y}^{t-1}=\mathcal{M}^{t-1}(X^t)$ and
$\overline{\widehat{Y}^{t-1}}=\mathcal{M}^{t-1}(\overline{X^t})$.}

The set of sentence-duplets with $\mathcal{D}^t$ and $\overline{\mathcal{D}^t}$ is denoted as:
\begin{equation}
\begin{aligned}
&(\mathcal{D}^t,\overline{\mathcal{D}^t}) = \left\{(X^t_i, \widetilde{\bm{Y}}^t_i, \overline{X^t_i}, \overline{\widetilde{\bm{Y}}^t_i})\right\}_{i=1}^{\left|\mathcal{D}^t\right|}\\
&s.t.~(X^t_i, \widetilde{\bm{Y}}^t_i) \in \mathcal{D}^t, (\overline{X^t_i}, \overline{\widetilde{\bm{Y}}^t_i}) \in \overline{\mathcal{D}^t}\\
\end{aligned}
\end{equation}

Using the sentence-duplets $(\mathcal{D}^t,\overline{\mathcal{D}^t})$ at the $t$-th step, our sentence-duplet learning scheme updates the new model $\mathcal{M}^t$ by optimizing the following loss function:
\begin{equation}
\begin{aligned}
\mathcal{L}_{\text{s-dup}}(X^t,\overline{X^t};\Theta^t)
=
&\mathcal{L}_{\text{rdp}}(X^t,\widetilde{Y}^t,\widehat{Y}^{t-1};\Theta^t)\\
&+
\mathcal{L}_{\text{rdp}}(\overline{X^t},\overline{\widetilde{Y}^t},\overline{\widehat{Y}^{t-1}};\Theta^t).
\end{aligned}
\label{eq:s_dup}
\end{equation}
which mirrors the form of Equation \eqref{gsc} applied to both the original sentence $X^t$ and its corresponding erased version $\overline{X^t}$. 
{In Equation \eqref{eq:s_dup}, all supervision signals are explicitly included as inputs of
$\mathcal{L}_{\text{rdp}}$. 
Therefore, the loss for the original sentence $X^t$ is
computed using $\widetilde{Y}^t$ and $\widehat{Y}^{t-1}$, while the loss for the erased
sentence $\overline{X^t}$ is computed using $\overline{\widetilde{Y}^t}$ and
$\overline{\widehat{Y}^{t-1}}$.}

\subsubsection{Contextual Consistency Loss}

To further mitigate the biased context, we introduce a contextual consistency loss $\mathcal{L}_{\text{ctxc}}(X^t,\overline{X^t};\Theta^t)$, which imposes a constraint on the old entity types' context within each sentence pair $(X^t,\overline{X^t})$, ensuring consistency across contexts.
For old-entity-type tokens, the context related to new entity types is present in the original sentence $X^t$ and removed in the corresponding erased sentence $\overline{X^t}$. 
To simplify, we denote $\mathcal{O}(X^t)$ to depict the positions $\{{p}^{{o}}_j\}^{\mathcal{O}(X^t)}_{j=1}$ of pseudo-labeled old-entity-type tokens within the sentence $X^t$.
To mitigate the impact of biased context between old and new entity types, the predictions of the updated model $\mathcal{M}^t$ on old-entity-type tokens with new-entity-type-related context should be consistent with those without such context.
Therefore, $\mathcal{L}_{\text{ctxc}}(X^t,\overline{X^t};\Theta^t)$ is formulated as follows:
 \begin{equation}
\mathcal{L}_{\text{ctxc}}(X^t,\overline{{X}^t};\Theta^t)
=
\sum_{p\in \mathcal{O}(X^t)}
\|\widehat{Y}^{t}_{p,:}-\overline{\widehat{Y}^{t}_{p,:}}\|_2^2,
\end{equation}
{where $\widehat{Y}^{t}=\mathcal{M}^t(X^t;\Theta^t)$ and
$\overline{\widehat{Y}^{t}}=\mathcal{M}^t(\overline{X^t};\Theta^t)$ are the current-model predictions
for the original and erased sentences, respectively. 
The notation
$\widehat{Y}^{t}_{p,:}$ denotes the predicted probability distribution of the $p$-th token.}

{Finally, the total objective function of our proposed TBCL method is formulated as follows:}
 \begin{equation}
\mathcal{L}_{\text{tbcl}}(\Theta^t)
=
\mathcal{L}_{\text{s-dup}}(X^t,\overline{{X}^t};\Theta^t)
+
\beta\mathcal{L}_{\text{ctxc}}(X^t,\overline{{X}^t};\Theta^t),
\end{equation}
where $\beta$ is a hyperparameter that balances the significance of the loss terms.

\section{Experimental Setup}

To ensure a fair comparison with previous SOTA INER methods, we adhere to the experimental setup specified in CFNER~\cite{zheng2022distilling} and RDP~\cite{zhang2023task}. This involves utilizing identical benchmark datasets, INER settings, competing baselines, assessment metrics, and foundational implementation details.

\subsection{Benckmark Datasets}

We assess our TBCL using three NER datasets, which are CoNLL2003 \cite{sang1837introduction}, I2B2 \cite{murphy2010serving}, and OntoNotes5 \cite{hovy2006ontonotes}. Table \ref{tab:dataset_statistics} provides a summary of the dataset statistics.

To divide the training set into separate slices representing various incremental learning steps, we employ the greedy sampling algorithm presented in CFNER~\cite{zheng2022distilling}. This method guarantees that samples from each entity type are predominantly concentrated in their respective slice, thereby more accurately reflecting real-world scenarios. Within each slice, we keep only the labels related to the entity types being learned and classify all other labels as non-entity types. For additional details on this greedy sampling algorithm, please refer to CFNER's~\cite{zheng2022distilling} Appendix B.

\begin{table}[t]
  \centering
  \caption{The summary for three NER datasets.}
  \resizebox{1.0\linewidth}{!}{
    \begin{tabular}{ccccc}
    \toprule
     \textbf{Dataset}     & \multicolumn{1}{l}{\# \textbf{Entity Type}} & \# \textbf{Sample}  & \multicolumn{2}{c}{\textbf{Entity Type Sequence (Alphabetical Order)}} \\
    \midrule
    CoNLL2003~\cite{sang1837introduction} & 4     & 21k    & \multicolumn{2}{l}{\begin{tabular}[1]{l}LOCATION, MISC, ORGANISATION, PERSON\end{tabular}} \\
    \midrule
    I2B2 \cite{murphy2010serving}  & 16    & 141k   & \multicolumn{2}{c}{\begin{tabular}[1]{l}AGE, CITY, COUNTRY, DATE, DOCTOR, HOSPITAL, \\  IDNUM, MEDICALRECORD, ORGANIZATION, \\PATIENT, PHONE, PROFESSION, STATE, STREET, \\USERNAME, ZIP\end{tabular}} \\
    \midrule
    OntoNotes5 \cite{hovy2006ontonotes} & 18    & 77k   & \multicolumn{2}{c}{\begin{tabular}[1]{l}CARDINAL, DATE, EVENT, FAC, GPE, LANGUAGE,\\ LAW, LOC, MONEY, NORP, ORDINAL, ORG,\\ PERCENT, PERSON, PRODUCT, QUANTITY, TIME,\\ WORK\_OF\_ART\end{tabular}} \\
    \bottomrule
    \end{tabular}
    }
  \label{tab:dataset_statistics}%
\end{table}%

\subsection{INER Settings}

In accordance with CFNER~\cite{zheng2022distilling} and RDP \cite{zhang2023task}, we examine two distinct INER scenarios for each dataset. 
In the first scenario, the base model (initial step) is trained with the same number of entity types as in subsequent incremental learning steps. 
In the second scenario, the base model is trained with half of all entity types. 
The first scenario presents a greater challenge, whereas the second more accurately reflects real-world situations, allowing models to acquire adequate knowledge before entering incremental learning.
Entity types are learned sequentially during training, specifically in alphabetical order, with each corresponding data slice utilized to train the models sequentially. {The base model is trained on FG (First Group) entity types, and each incremental learning step introduces PG (Progressive Group) entity types, denoted as FG-a-PG-b.} 
For the CoNLL2003~\cite{sang1837introduction} dataset, two settings are employed: FG-1-PG-1 and FG-2-PG-1. For the I2B2~\cite{murphy2010serving} and OntoNotes5~\cite{hovy2006ontonotes} datasets, four settings are established: FG-1-PG-1, FG-2-PG-2, FG-8-PG-1, and FG-8-PG-2.
During evaluation, only the labels pertinent to the current entity types in the validation set are retained, while all others are classified as non-entity types. In each incremental learning step, the model with the best validation performance is selected for testing and for the subsequent incremental learning phase. When testing, labels for all previously learned entity types are preserved in the test set, while others are marked as non-entity types.

\subsection{Competing Baselines}

We incorporate the following baselines, including SOTA INER methods: ExtendNER~\cite{monaikul2021continual}, CFNER~\cite{zheng2022distilling}, and RDP~\cite{zhang2023task}. Additionally, we include the lower-bound method, Only Finetuning, which uses new data directly to finetune the model without employing any anti-forgetting techniques. 
Furthermore, we integrate incremental learning methods adapted from computer vision, such as PODNet~\cite{DBLP:conf/eccv/DouillardCORV20}, LUCIR~\cite{DBLP:conf/cvpr/HouPLWL19}, and Self-Training~\cite{DBLP:journals/corr/abs-1909-08383}, which are tailored to the INER scenario~\cite{zheng2022distilling}. 
Detailed introductions to these baselines are provided below:

\begin{itemize}

\item PODNet \cite{DBLP:conf/eccv/DouillardCORV20}: Originally designed to tackle the issue of catastrophic forgetting in incremental learning for image classification, this method has been adapted to the INER scenario \cite{zheng2022distilling}. The model's total loss is composed of classification and distillation losses. For classification, instead of the standard cross-entropy loss, PODNet utilizes neighborhood component analysis loss. In terms of distillation loss, PODNet enforces constraints on the output of each intermediate layer.

\item LUCIR \cite{DBLP:conf/cvpr/HouPLWL19}: This method creates an incremental learning framework for image classification tasks, which is then transferred to the INER scenario \cite{zheng2022distilling}. Similar to PODNet, it aims to minimize catastrophic forgetting. Its loss function consists of three main components: (1) cross-entropy loss for samples containing new entity types; (2) distillation loss between features extracted by the previous and current models; and (3) margin-ranking loss for the retained samples with the previous entity types.

\item Self-Training \cite{DBLP:conf/wacv/RosenbergHS05,DBLP:journals/corr/abs-1909-08383}: This method directly employs the pre-existing model to label the non-entity type tokens with their previous entity types. The updated model is then trained using new data that encompasses annotations for all previously identified entity types. Its goal is to minimize cross-entropy loss on these data, ensuring efficient model training.

\item ExtendNER \cite{monaikul2021continual}: Similar to Self-Training, ExtendNER applies KD to the INER. It computes cross-entropy loss for entity type tokens and KL divergence loss for non-entity tokens. During training, ExtendNER aims to minimize the combined losses of cross-entropy and KL divergence.

\item CFNER \cite{zheng2022distilling}: Building on ExtendNER, this method introduces a causal framework for INER that distills causal effects within the non-entity type tokens. Initially, the old model identifies the non-entity type tokens associated with previous entity types for KD, while curriculum learning helps reduce recognition errors.

\item RDP \cite{zhang2023task}: This method is a pseudo-labeling based INER method, targeting catastrophic forgetting and the semantic shift issue of the non-entity type tokens within INER. It incorporates a task relation distillation loss to achieve a balance between stability and adaptability, and utilizes a prototypical pseudo-label strategy to reduce label noise and address the issue of semantic shift.
\end{itemize}

\subsection{Assessment Metrics}
At each incremental learning step, we first compute the F1 score for each entity type, then calculate the macro-average F1 scores (Macro-F1) for old entity types, new entity types, and all entity types at the current step. We also compute the micro-average F1 score (Micro-F1) for all current entity types. We provide the average scores over all steps, including the first one (except for old and new entity types). Additionally, we provide line plots showing step-wise changes in scores.
To check the statistical significance of the improvements, we apply a paired t-test at a significance level of $0.05$.

\begin{table*}[t]
\centering
\caption{Comparisons with baselines on the CoNLL2003~\cite{sang1837introduction} dataset. The \textcolor{deepred}{\textbf{red}} denotes the highest result, while the \textcolor{blue}{\textbf{blue}} denotes the second highest result. The marker $\dagger$ refers to significant test $p$-$value<0.05$ comparing with RDP~\cite{zhang2023task}. $*$ represents results from our re-implementation. Other baseline results are directly referenced from RDP~\cite{zhang2023task}.}
\resizebox{1.0\linewidth}{!}{%
\begin{tabular}{@{}ccccccccc@{}}
\toprule
&\multicolumn{4}{c}{FG-1-PG-1} & \multicolumn{4}{c}{FG-2-PG-1}  \\ \cmidrule(l){2-5} \cmidrule(l){6-9} 
&\multicolumn{1}{c}{Old Entity Types} & \multicolumn{1}{c}{New Entity Types}&\multicolumn{2}{c}{All Entity Types}& \multicolumn{1}{c}{Old Entity Types} & \multicolumn{1}{c}{New Entity Types}&\multicolumn{2}{c}{All Entity Types} \\
\cmidrule(l){4-5} \cmidrule(l){8-9}
 \multirow{-4}{*}{Baseline} & Avg. Macro-F1   & Avg. Macro-F1  & Avg. Micro-F1  & Avg. Macro-F1  & Avg. Macro-F1   & Avg. Macro-F1  &Avg. Micro-F1  &Avg. Macro-F1  \\ \midrule
 Only Finetuning &-- &-- & 50.84±0.10 & 40.64±0.16 &-- &-- & 57.45±0.05 & 43.58±0.18 \\
 
 PODNet~\cite{DBLP:conf/eccv/DouillardCORV20} &-- &-- & 36.74±0.52 & 29.43±0.28 &-- &-- & 59.12±0.54 & 58.39±0.99 \\

 LUCIR~\cite{DBLP:conf/cvpr/HouPLWL19} &-- &-- & 74.15±0.43 & 70.48±0.66 &-- &-- & 80.53±0.31 & 77.33±0.31 \\

 Self-Training~\cite{DBLP:conf/wacv/RosenbergHS05} & --& --& 76.17±0.91 & 72.88±1.12 &-- & --& 76.65±0.24 & 66.72±0.11 \\

  $\text{ExtendNER}^{*}$~\cite{monaikul2021continual} & 70.42±0.22 & 71.84±0.69 & 76.07±0.35 & 73.06±0.29 & 56.33±2.00 & 79.34±0.71 & 77.89±0.42 & 69.92±1.02 \\

 ExtendNER~\cite{monaikul2021continual} &-- & --& 76.36±0.98 & 73.04±1.80 &-- &-- & 76.66±0.66 & 66.36±0.64 \\

 $\text{CFNER}^{*}$~\cite{zheng2022distilling} & 76.82±0.58 & 78.87±0.44 & 80.29±0.21 & 78.44±0.24 & 69.94±1.63 & 84.08±0.26 & 81.52±0.43 & 77.20±0.82  \\

 CFNER~\cite{zheng2022distilling} &-- &-- & 80.91±0.29 & 79.11±0.50 &-- &-- & 80.83±0.36 & 75.20±0.32  \\

 RDP \cite{zhang2023task} & \textcolor{blue}{\textbf{79.87±0.37}} &\textcolor{blue}{\textbf{82.52±0.32}} & \textcolor{blue}{\textbf{82.55±0.26}} & \textcolor{blue}{\textbf{80.64±0.12}} & \textcolor{blue}{\textbf{82.44±0.73}} & \textcolor{blue}{\textbf{88.34±0.35}} & \textcolor{blue}{\textbf{85.32±0.36}} & \textcolor{blue}{\textbf{83.59±0.37}} \\

 \midrule

 \textbf{TBCL (Ours)} & $\textcolor{deepred}{\textbf{80.95±0.28}}^{\dagger}$ & $\textcolor{deepred}{\textbf{83.68±0.19}}^{\dagger}$ & $\textcolor{deepred}{\textbf{84.27±0.38}}^{\dagger}$ & $\textcolor{deepred}{\textbf{81.92±0.16}}^{\dagger}$ & $\textcolor{deepred}{\textbf{83.76±0.66}}^{\dagger}$ & $\textcolor{deepred}{\textbf{89.58±0.47}}^{\dagger}$ & $\textcolor{deepred}{\textbf{86.71±0.56}}^{\dagger}$ & $\textcolor{deepred}{\textbf{84.94±0.53}}^{\dagger}$ \\
 
\textbf{Improve} & $\Uparrow$\textbf{1.08} & $\Uparrow$\textbf{1.16}& $\Uparrow$\textbf{1.72}& $\Uparrow$\textbf{1.28}& $\Uparrow$\textbf{1.32}& $\Uparrow$\textbf{1.24}& $\Uparrow$\textbf{1.39}& $\Uparrow$\textbf{1.35} \\

\bottomrule
\end{tabular}
}
\label{tab:main_result_1_full}
\end{table*}

\begin{table*}[t]
\centering
\caption{Comparisons with baselines on the I2B2~\cite{murphy2010serving} dataset. The \textcolor{deepred}{\textbf{red}} denotes the highest result, while the \textcolor{blue}{\textbf{blue}} denotes the second highest result. The marker $\dagger$ refers to significant test $p$-$value<0.05$ comparing with RDP~\cite{zhang2023task}. $*$ represents results from our re-implementation. Other baseline results are directly referenced from RDP~\cite{zhang2023task}.}
\resizebox{1.0\linewidth}{!}{%
\begin{tabular}{@{}ccccccccc@{}}
\toprule
&\multicolumn{4}{c}{FG-1-PG-1} & \multicolumn{4}{c}{FG-2-PG-2}  \\ \cmidrule(l){2-5} \cmidrule(l){6-9} 
&\multicolumn{1}{c}{Old Entity Types} & \multicolumn{1}{c}{New Entity Types}&\multicolumn{2}{c}{All Entity Types}& \multicolumn{1}{c}{Old Entity Types} & \multicolumn{1}{c}{New Entity Types}&\multicolumn{2}{c}{All Entity Types} \\
\cmidrule(l){4-5} \cmidrule(l){8-9}
 \multirow{-4}{*}{Baseline} & Avg. Macro-F1   & Avg. Macro-F1  & Avg. Micro-F1  & Avg. Macro-F1  & Avg. Macro-F1   & Avg. Macro-F1  &Avg. Micro-F1  &Avg. Macro-F1  \\\midrule
 Only Finetuning &-- &-- & 17.43±0.54 & 13.81±1.14 & --&-- & 28.57±0.26 & 21.43±0.41 \\
 
 PODNet~\cite{DBLP:conf/eccv/DouillardCORV20} &-- &-- & 12.31±0.35 & 17.14±1.03 &-- &-- & 34.67±2.65 & 24.62±1.76 \\

 LUCIR~\cite{DBLP:conf/cvpr/HouPLWL19} &-- &-- & 43.86±2.43 & 31.31±1.62 &-- & --& 64.32±0.76 & 43.53±0.59 \\

 Self-Training~\cite{DBLP:conf/wacv/RosenbergHS05} &-- &-- & 31.98±2.12 & 14.76±1.31 &-- &-- & 55.44±4.78 & 33.38±3.13\\

  $\text{ExtendNER}^{*}$~\cite{monaikul2021continual} & 15.67±2.47 & 30.01±5.07 & 41.65±10.11 & 23.11±2.70 & 35.66±2.26 & 50.78±1.17 & 67.60±1.15 & 42.58±1.59 \\

 ExtendNER~\cite{monaikul2021continual} &-- &-- & 42.85±2.86 & 24.05±1.35 &-- &-- & 57.01±4.14 & 35.29±3.38 \\

 $\text{CFNER}^{*}$~\cite{zheng2022distilling} & 34.35±1.00 & 45.70±1.23 & 64.79±0.26 & 37.79±0.65 & 48.14±1.10 & 56.56±1.08 & 72.58±0.59 & 51.71±0.84  \\

 CFNER~\cite{zheng2022distilling} &-- &-- & 62.73±3.62 & 36.26±2.24 &-- & --& 71.98±0.50 & 49.09±1.38  \\

 RDP \cite{zhang2023task} & \textcolor{blue}{\textbf{41.61±2.71}} & \textcolor{blue}{\textbf{50.32±1.02}} & \textcolor{blue}{\textbf{71.39±1.01}} & \textcolor{blue}{\textbf{44.00±2.31}} & \textcolor{blue}{\textbf{49.07±1.01}} & \textcolor{blue}{\textbf{62.13±0.77}} & \textcolor{blue}{\textbf{77.45±0.55}} & \textcolor{blue}{\textbf{53.48±0.66}} \\

 \midrule

 \textbf{TBCL (Ours)} & $\textcolor{deepred}{\textbf{44.79±1.78}}^{\dagger}$ & $\textcolor{deepred}{\textbf{52.08±0.42}}^{\dagger}$ & $\textcolor{deepred}{\textbf{72.14±0.75}}^{\dagger}$ & $\textcolor{deepred}{\textbf{48.66±1.40}}^{\dagger}$ & $\textcolor{deepred}{\textbf{49.62±1.37}}^{\dagger}$ & $\textcolor{deepred}{\textbf{62.85±0.71}}^{\dagger}$ & $\textcolor{deepred}{\textbf{78.19±0.26}}^{\dagger}$ & $\textcolor{deepred}{\textbf{54.46±0.63}}^{\dagger}$ \\
 
\textbf{Improve} & $\Uparrow$\textbf{3.18}& $\Uparrow$\textbf{1.76}& $\Uparrow$\textbf{0.75}& $\Uparrow$\textbf{4.66}& $\Uparrow$\textbf{0.55}& $\Uparrow$\textbf{0.72}& $\Uparrow$\textbf{0.74}& $\Uparrow$\textbf{0.98} \\
\midrule
&\multicolumn{4}{c}{FG-8-PG-1} & \multicolumn{4}{c}{FG-8-PG-2}  \\ \cmidrule(l){2-5} \cmidrule(l){6-9} 
&\multicolumn{1}{c}{Old Entity Types} & \multicolumn{1}{c}{New Entity Types}&\multicolumn{2}{c}{All Entity Types}& \multicolumn{1}{c}{Old Entity Types} & \multicolumn{1}{c}{New Entity Types}&\multicolumn{2}{c}{All Entity Types} \\
\cmidrule(l){4-5} \cmidrule(l){8-9}
 \multirow{-4}{*}{Baseline} & Avg. Macro-F1   & Avg. Macro-F1  & Avg. Micro-F1  & Avg. Macro-F1  & Avg. Macro-F1   & Avg. Macro-F1  &Avg. Micro-F1  &Avg. Macro-F1  \\\midrule
 Only Finetuning &-- &-- & 20.83±1.78 & 18.11±1.66 &-- &-- & 23.60±0.15 & 23.54±0.38 \\
 
 PODNet~\cite{DBLP:conf/eccv/DouillardCORV20} &-- &-- & 39.26±1.38 & 27.23±0.93 &-- &-- & 36.22±12.9 & 26.08±7.42 \\

 LUCIR~\cite{DBLP:conf/cvpr/HouPLWL19} &-- &-- & 57.86±0.87& 33.04±0.39 &-- &-- & 68.54±0.27 & 46.94±0.63 \\

 Self-Training~\cite{DBLP:conf/wacv/RosenbergHS05} &-- &-- & 49.51±1.35 & 23.77±1.01 &-- &-- & 48.94±6.78 & 29.00±3.04\\

  $\text{ExtendNER}^{*}$~\cite{monaikul2021continual} & 20.33±0.99 & 27.65±1.30 & 45.14±2.91 & 27.41±0.88 & 27.01±2.10 & 39.22±1.19 & 56.48±2.41 & 38.88±1.38 \\

 ExtendNER~\cite{monaikul2021continual} &-- &-- & 43.95±2.01 & 23.12±1.79 &-- &-- & 52.25±5.36 & 30.93±2.77 \\

 $\text{CFNER}^{*}$~\cite{zheng2022distilling} & 31.13±1.56 & 37.95±2.08 & 56.66±3.22 & 36.84±1.35 & 43.94±1.25 & 49.93±1.19 & 69.12±0.94 & 51.61±0.87 \\

 CFNER~\cite{zheng2022distilling} &-- &-- & 59.79±1.70 & 37.30±1.15 &-- &-- & 69.07±0.89 & 51.09±1.05 \\

 RDP \cite{zhang2023task} & \textcolor{blue}{\textbf{61.75±0.36}} &\textcolor{blue}{\textbf{52.47±1.20}} & \textcolor{blue}{\textbf{77.50±1.26}} & \textcolor{blue}{\textbf{62.99±0.36}} &\textcolor{blue}{\textbf{60.44±0.92}} &\textcolor{blue}{\textbf{56.17±1.17}} & \textcolor{blue}{\textbf{80.08±0.40}} & \textcolor{blue}{\textbf{63.72±0.71}} \\

 \midrule

 \textbf{TBCL (Ours)} & $\textcolor{deepred}{\textbf{66.53±0.47}}^{\dagger}$ & $\textcolor{deepred}{\textbf{58.13±0.41}}^{\dagger}$ & $\textcolor{deepred}{\textbf{83.99±0.29}}^{\dagger}$ & $\textcolor{deepred}{\textbf{67.27±0.28}}^{\dagger}$ & $\textcolor{deepred}{\textbf{62.49±1.08}}^{\dagger}$ & $\textcolor{deepred}{\textbf{58.96±0.51}}^{\dagger}$ & $\textcolor{deepred}{\textbf{82.37±1.19}}^{\dagger}$ & $\textcolor{deepred}{\textbf{66.51±1.32}}^{\dagger}$ \\
 
\textbf{Improve} & $\Uparrow$\textbf{4.78}& $\Uparrow$\textbf{5.66}& $\Uparrow$\textbf{6.49}& $\Uparrow$\textbf{4.28}& $\Uparrow$\textbf{2.05}& $\Uparrow$\textbf{2.79}& $\Uparrow$\textbf{2.29}& $\Uparrow$\textbf{2.79} \\

\bottomrule
\end{tabular}
}
\label{tab:main_result_2_full}
\end{table*}

\begin{table*}[t]
\centering
\caption{Comparisons with baselines on the OntoNotes5~\cite{hovy2006ontonotes} dataset. The \textcolor{deepred}{\textbf{red}} denotes the highest result, while the \textcolor{blue}{\textbf{blue}} denotes the second highest result. The marker $\dagger$ refers to significant test $p$-$value<0.05$ comparing with RDP~\cite{zhang2023task}. $*$ represents results from our re-implementation. Other baseline results are directly referenced from RDP~\cite{zhang2023task}.}
\resizebox{1.0\linewidth}{!}{%
\begin{tabular}{@{}ccccccccc@{}}
\toprule
&\multicolumn{4}{c}{FG-1-PG-1} & \multicolumn{4}{c}{FG-2-PG-2}  \\ \cmidrule(l){2-5} \cmidrule(l){6-9} 
&\multicolumn{1}{c}{Old Entity Types} & \multicolumn{1}{c}{New Entity Types}&\multicolumn{2}{c}{All Entity Types}& \multicolumn{1}{c}{Old Entity Types} & \multicolumn{1}{c}{New Entity Types}&\multicolumn{2}{c}{All Entity Types} \\
\cmidrule(l){4-5} \cmidrule(l){8-9}
 \multirow{-4}{*}{Baseline} & Avg. Macro-F1   & Avg. Macro-F1  & Avg. Micro-F1  & Avg. Macro-F1  & Avg. Macro-F1   & Avg. Macro-F1  &Avg. Micro-F1  &Avg. Macro-F1  \\\midrule
 Only Finetuning &-- &-- & 15.27±0.26 & 10.86±1.11 &-- & --&25.85±0.11 & 20.55±0.24 \\
 
 PODNet~\cite{DBLP:conf/eccv/DouillardCORV20} &-- &-- & 9.06±0.56 & 8.36±0.57 &-- &-- & 34.67±1.08 & 24.62±0.85 \\

 LUCIR~\cite{DBLP:conf/cvpr/HouPLWL19} &-- &-- & 28.18±1.15 & 21.11±0.84 &-- &-- & 64.32±1.79 & 43.53±1.11 \\

 Self-Training~\cite{DBLP:conf/wacv/RosenbergHS05} &-- &-- & 50.71±0.79 & 33.24±1.06 &-- &-- & 68.93±1.67 & 50.63±1.66\\

  $\text{ExtendNER}^{*}$~\cite{monaikul2021continual} & 29.63±1.14 &41.11±1.07 & 51.36±0.77 & 33.38±0.98 & 44.90±6.49 & 51.97±2.87 & 63.03±9.39 & 47.64±5.15 \\

 ExtendNER~\cite{monaikul2021continual} &-- &-- & 50.53±0.86 & 32.84±0.84 &-- &-- & 67.61±1.53 & 49.26±1.49 \\

 $\text{CFNER}^{*}$~\cite{zheng2022distilling} & 37.74±1.77 & 55.01±0.20 & 58.44±0.71 & 41.75±1.51 & 52.70±0.26 & 60.54±0.99 & 72.10±0.31 & 55.02±0.35  \\

 CFNER~\cite{zheng2022distilling} &-- &-- & 58.94±0.57 & 42.22±1.10 &-- &-- & 72.59±0.48 & 55.96±0.69 \\

 RDP \cite{zhang2023task} & \textcolor{blue}{\textbf{52.71±0.48}} &\textcolor{blue}{\textbf{59.66±0.34}} & \textcolor{blue}{\textbf{68.28±1.09}} & \textcolor{blue}{\textbf{53.56±0.39}} &\textcolor{blue}{\textbf{55.77±0.71}} &\textcolor{blue}{\textbf{63.93±0.47}} & \textcolor{blue}{\textbf{74.38±0.26}} & \textcolor{blue}{\textbf{57.73±0.54}} \\

 \midrule

 \textbf{TBCL (Ours)} &$\textcolor{deepred}{\textbf{54.36±0.63}}^{\dagger}$ &$\textcolor{deepred}{\textbf{61.53±0.67}}^{\dagger}$ & $\textcolor{deepred}{\textbf{70.04±0.73}}^{\dagger}$ & $\textcolor{deepred}{\textbf{55.63±0.51}}^{\dagger}$ &$\textcolor{deepred}{\textbf{57.70±0.69}}^{\dagger}$ &$\textcolor{deepred}{\textbf{66.09±0.59}}^{\dagger}$ & $\textcolor{deepred}{\textbf{76.40±0.44}}^{\dagger}$ & $\textcolor{deepred}{\textbf{59.84±0.53}}^{\dagger}$ \\
 
\textbf{Improve} & $\Uparrow$\textbf{1.65}& $\Uparrow$\textbf{1.87}& $\Uparrow$\textbf{1.76}& $\Uparrow$\textbf{2.07}& $\Uparrow$\textbf{1.93}& $\Uparrow$\textbf{2.16}& $\Uparrow$\textbf{2.02}& $\Uparrow$\textbf{2.11} \\

\midrule
&\multicolumn{4}{c}{FG-8-PG-1} & \multicolumn{4}{c}{FG-8-PG-2}  \\ \cmidrule(l){2-5} \cmidrule(l){6-9} 
&\multicolumn{1}{c}{Old Entity Types} & \multicolumn{1}{c}{New Entity Types}&\multicolumn{2}{c}{All Entity Types}& \multicolumn{1}{c}{Old Entity Types} & \multicolumn{1}{c}{New Entity Types}&\multicolumn{2}{c}{All Entity Types} \\
\cmidrule(l){4-5} \cmidrule(l){8-9}
 \multirow{-4}{*}{Baseline} & Avg. Macro-F1   & Avg. Macro-F1  & Avg. Micro-F1  & Avg. Macro-F1  & Avg. Macro-F1   & Avg. Macro-F1  &Avg. Micro-F1  &Avg. Macro-F1  \\\midrule
 Only Finetuning &-- &-- & 17.63±0.57 & 12.23±1.08 &-- &-- & 29.81±0.12 & 20.05±0.16 \\
 
 PODNet~\cite{DBLP:conf/eccv/DouillardCORV20} &-- &-- & 29.00±0.86 & 20.54±0.91 &-- &-- & 37.38±0.26 & 25.85±0.29 \\

 LUCIR~\cite{DBLP:conf/cvpr/HouPLWL19} &-- &-- & 66.46±0.46 & 46.29±0.38 &-- &-- & 76.17±0.09 & 55.58±0.55 \\

 Self-Training~\cite{DBLP:conf/wacv/RosenbergHS05} &-- &-- & 73.59±0.66 & 49.41±0.77 &-- &-- & 77.07±0.62 & 53.32±0.63\\

  $\text{ExtendNER}^{*}$~\cite{monaikul2021continual} & 48.73±0.63 & 46.09±0.56 & 73.65±0.19 & 50.55±0.56 & 51.11±0.70 & 57.41±0.67 & 77.86±0.10 & 55.21±0.51 \\

 ExtendNER~\cite{monaikul2021continual} &-- &-- & 73.12±0.93 & 49.55±0.90 &-- &-- & 76.85±0.77 & 54.37±0.57 \\

 $\text{CFNER}^{*}$~\cite{zheng2022distilling} & 57.28±0.49 & 59.73±0.60 & 78.25±0.33 & 58.64±0.42 & 58.03±0.47 & 65.01±0.72 & 80.09±0.37 & 61.06±0.37  \\

 CFNER~\cite{zheng2022distilling} & -- & -- & 78.92±0.58 & 57.51±1.32 &-- &-- & 80.68±0.25 & 60.52±0.84  \\

 RDP \cite{zhang2023task} &\textcolor{blue}{\textbf{62.10±6.44}} &\textcolor{blue}{\textbf{62.58±3.44}} & \textcolor{blue}{\textbf{79.89±0.20}} & \textcolor{blue}{\textbf{63.20±0.58}} &\textcolor{blue}{\textbf{65.25±1.59}} &\textcolor{blue}{\textbf{67.10±1.30}} & \textcolor{blue}{\textbf{83.30±0.30}} & \textcolor{blue}{\textbf{66.92±1.26}} \\

 \midrule

 \textbf{TBCL (Ours)} & $\textcolor{deepred}{\textbf{65.47±0.33}}^{\dagger}$ & $\textcolor{deepred}{\textbf{64.01±0.42}}^{\dagger}$ & $\textcolor{deepred}{\textbf{81.41±0.16}}^{\dagger}$ & $\textcolor{deepred}{\textbf{66.51±0.28}}^{\dagger}$ & $\textcolor{deepred}{\textbf{66.40±0.49}}^{\dagger}$ & $\textcolor{deepred}{\textbf{69.22±0.85}}^{\dagger}$ & $\textcolor{deepred}{\textbf{83.80±0.21}}^{\dagger}$ & $\textcolor{deepred}{\textbf{68.43±0.41}}^{\dagger}$ \\
 
\textbf{Improve} & $\Uparrow$\textbf{3.37}& $\Uparrow$\textbf{1.43}& $\Uparrow$\textbf{1.52}& $\Uparrow$\textbf{3.31}& $\Uparrow$\textbf{1.15}& $\Uparrow$\textbf{2.12}& $\Uparrow$\textbf{0.50}& $\Uparrow$\textbf{1.51} \\

\bottomrule
\end{tabular}
}
\label{tab:main_result_3_full}
\end{table*}

\subsection{Implementation Details}

In line with previous INER methods~\cite{monaikul2021continual, xia2022learn, zheng2022distilling, zhang2023task}, we utilize the ``BIO'' tagging scheme for all three datasets. This scheme designates two labels for each entity type: B-type (\emph{i.e.}, for the beginning of an entity) and I-type (\emph{i.e.}, for the interior of an entity). Our NER model employs bert-base-cased \cite{kenton2019bert} as the encoder, complemented by a fully connected layer for classification. The model is implemented using the PyTorch framework \cite{paszke2019pytorch} and is built upon the BERT implementation from Huggingface \cite{wolf2019huggingface}.
As per CFNER~\cite{zheng2022distilling} and RDP~\cite{zhang2023task}, the model is trained for $20$ epochs if PG is set to $2$, and for $10$ epochs in other cases. The batch size, learning rate, and balancing hyper-parameter $\gamma$ are set to $8$, $4e-4$, and $0.01$, respectively.
All experiments are run on a single NVIDIA A100 GPU with 40GB of memory, and each experiment is repeated \textbf{five times} to ensure statistical reliability.

\section{Experimental Results}

To showcase the superiority and effectiveness of our TBCL method, we conducted extensive experiments focusing on the following research questions (RQ):

\begin{itemize}
\item \textbf{RQ1}: How does the quantitative performance of TBCL compare to that of competitive baselines?
\item \textbf{RQ2}: What effects do the key components of TBCL have?
\item \textbf{RQ3}: How does the qualitative performance of TBCL compare to that of competitive baselines?

\item \textbf{RQ4}: {How does TBCL maintain stability across different incremental scenarios, including 1) varying entity type learning orders, 2) employing larger language models as the encoder backbone, and 3) handling finer-grained entity types?}
\end{itemize}

\subsection{Main Results (RQ1)}

In this subsection, we performed extensive experiments on the CoNLL2003 \cite{sang1837introduction}, I2B2 \cite{murphy2010serving}, and OntoNotes5 \cite{hovy2006ontonotes} datasets under ten INER settings to showcase the quantitative performance of our TBCL method compared to seven competitive baselines. Tables \ref{tab:main_result_1_full}, \ref{tab:main_result_2_full}, and \ref{tab:main_result_3_full} present the average Macro-F1 scores for the old, new, and all entity types across all INER steps for each setting, along with the average Micro-F1 score for all entity types across all INER steps. Each experiment was conducted five times to ensure statistical robustness.

Specifically, as shown in Table \ref{tab:main_result_1_full}, for the FG-1-PG-1 and FG-2-PG-1 settings of the CoNLL dataset \cite{sang1837introduction}, our TBCL method achieved average improvements of $1.72$ and $1.39$ in Micro-F1 scores, and $1.28$ and $1.35$ in Macro-F1 scores, respectively, for all entity types at each step compared to the previous SOTA RDP method. For old entity types at each step, our TBCL method achieved average improvements of $1.08$ and $1.32$ in Macro-F1 scores, respectively. For new entity types at each step, our TBCL method achieved average improvements of $1.16$ and $1.24$ in Macro-F1 scores, respectively.

As shown in Table \ref{tab:main_result_2_full}, for the FG-1-PG-1, FG-2-PG-2, FG-8-PG-1, and FG-8-PG-2 settings of the I2B2 dataset \cite{murphy2010serving}, our TBCL method achieved average improvements of $0.75$, $0.74$, $6.49$, and $2.29$ in Micro-F1 scores, and $4.66$, $0.98$, $4.28$, and $2.79$ in Macro-F1 scores, respectively, for all entity types at each step compared to the previous SOTA RDP method. For old entity types at each step, our TBCL method achieved average improvements of $3.18$, $0.55$, $4.78$, and $2.05$ in Macro-F1 scores, respectively. For new entity types at each step, our TBCL method achieved average improvements of $1.76$, $0.72$, $5.66$, and $2.79$ in Macro-F1 scores, respectively.

As shown in Table \ref{tab:main_result_3_full}, for the FG-1-PG-1, FG-2-PG-2, FG-8-PG-1, and FG-8-PG-2 settings of the OntoNotes5 dataset \cite{hovy2006ontonotes}, our TBCL method achieved average improvements of $1.76$, $2.02$, $1.52$, and $0.50$ in Micro-F1 scores, and $2.07$, $2.11$, $3.31$, and $1.51$ in Macro-F1 scores, respectively, for all entity types at each step compared to the previous SOTA RDP method. For old entity types at each step, our TBCL method achieved average improvements of $1.65$, $1.93$, $3.37$, and $1.15$ in Macro-F1 scores, respectively. For new entity types at each step, our TBCL method achieved average improvements of $1.87$, $2.16$, $1.43$, and $2.12$ in Macro-F1 scores, respectively.

These results quantitatively validate the superiority and effectiveness of our TBCL method over competitive baselines, highlighting its capability to learn a robust INER model. The findings suggest enhanced resilience to catastrophic forgetting and semantic shift problems. Moreover, our TBCL method further enhances the performance of the pseudo-labeling based RDP method by rectifying biased context, thereby mitigating old-entity-type forgetting and new-entity-type overfitting.

{To provide a more detailed discussion of these numerical results, we analyzed TBCL's performance across CoNLL2003, I2B2, and OntoNotes5 datasets (Tables \ref{tab:main_result_1_full}--\ref{tab:main_result_3_full}). 
On CoNLL2003, FG-1-PG-1 and FG-2-PG-1 yield modest Micro-F1 gains of 1.39–1.72 and Macro-F1 gains of 1.28–1.35, as the dataset is simple and RDP performs near its upper bound. 
In I2B2, the largest Micro-F1 improvements are observed in FG-8-PG-1 (+6.49) and FG-8-PG-2 (+2.29), while FG-1-PG-1 and FG-2-PG-2 show smaller gains, reflecting the higher potential for improvement when the initial number of types is larger. 
OntoNotes5 shows similar trends, with FG-2-PG-2 having the largest Micro-F1 gain (+2.02) and FG-8-PG-1 the largest Macro-F1 gain (+3.31). 
These results indicate that TBCL is particularly effective in challenging, multi-entity-type incremental learning settings, while still providing consistent improvements in simpler settings. 
Overall, TBCL demonstrates both incremental gains and substantial methodological advancements, emphasizing its practical and scientific contributions.}

\begin{figure}[tbp]
\centering
  \includegraphics[width=1.0\linewidth]{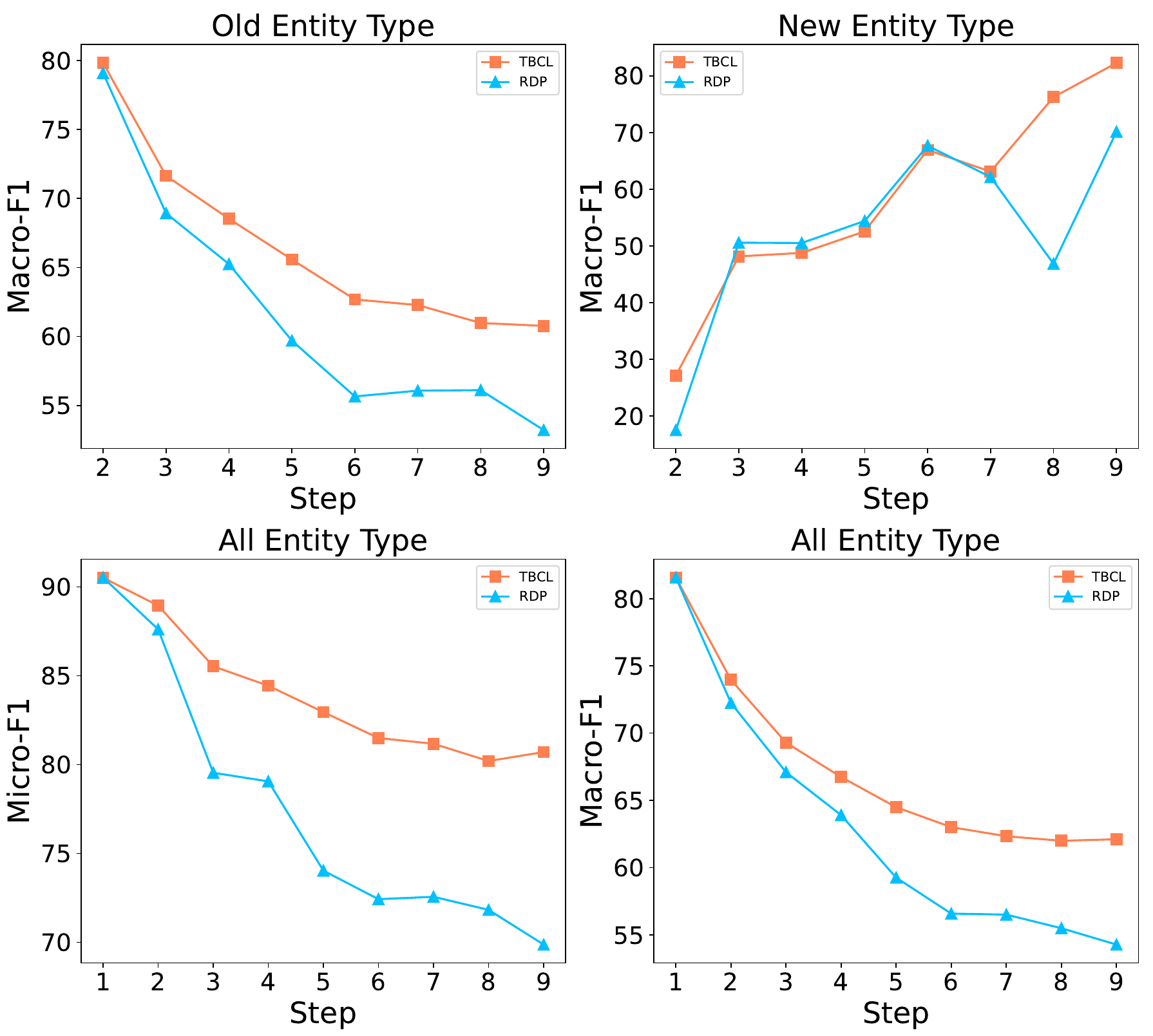}
    \caption{Comparison of the step-wise Macro-F1 scores for old, new, and all entity types, along with the step-wise Micro-F1 scores for all entity types, using the FG-8-PG-1 setting of the I2B2 dataset \cite{murphy2010serving}. Our TBCL method consistently outperforms the previous SOTA INER baseline, RDP \cite{zhang2023task}, in most step-wise evaluations.}
    \label{fig:step}
\end{figure}

\subsection{Step-wise Results (RQ1)}

As illustrated in Figure \ref{fig:step}, we provide a detailed step-wise comparison to thoroughly assess the effectiveness of our TBCL method in handling INER scenarios. The results clearly show that TBCL consistently outperforms the previous SOTA INER model, RDP \cite{zhang2023task}, in most step-wise evaluations. This is evident in the FG-8-PG-1 setting of the I2B2 dataset \cite{murphy2010serving}. These findings highlight the robustness and adaptability of TBCL, as it effectively maintains high performance when encountering new and evolving entity types throughout the learning process. The substantial improvement over RDP suggests that TBCL has the potential to set a new benchmark for INER tasks, particularly in complex, real-world data environments.

\subsection{Ablation Study (RQ2)}

In this subsection, we begin with ablation experiments to validate the effectiveness of the sentence-duplet learning scheme. Following this, we proceed to validate the contextual consistency loss through further experiments. 
All ablation experiments are conducted under the FG-1-PG-1 setting of the I2B2 dataset \cite{murphy2010serving}. Our baseline is established from the SOTA pseudo-labeling based INER method RDP \cite{zhang2023task}. 

\begin{table}[tbp]
  \centering
  \caption{The ablation study of our TBCL under the FG-1-PG-1 setting of the I2B2 dataset \cite{murphy2010serving}. In comparison to our TBCL method, all ablation variants show a significant decline in INER performance, confirming the necessity of each component for collaboratively addressing INER. \textbf{Bold} denotes the best results.}
  \resizebox{1.0\linewidth}{!}{
    \begin{tabular}{lcccc}
    \toprule
    \multirow{2}[4]{*}{Method} & {Old Entity Types} & {New Entity Types}&  \multicolumn{2}{c}{All Entity Types} \\
\cmidrule{4-5} & Avg. Macro-F1 & Avg. Macro-F1 & Avg. Micro-F1 & Avg. Macro-F1  \\
    \midrule
Baseline (RDP \cite{zhang2023task}) &41.61±2.71 &50.32±1.02 &71.39±1.01 &44.00±2.31  \\

Baseline\ +\ double &39.77±2.03 &47.47±1.13 &70.26±1.39 &42.52±1.66  \\

Baseline\ +\ duplet &43.63±1.48 &51.55±1.05 &71.88±0.89 &47.39±1.28  \\
    
  \midrule
Baseline\ +\ duplet &43.63±1.48 &51.55±1.05 &71.88±0.89 &47.39±1.28  \\

Baseline\ +\ duplet\ +\ ctxc (\textbf{TBCL}) &\textbf{44.79±1.78} &\textbf{52.08±0.42} & \textbf{72.14±0.75} & \textbf{48.66±1.40}\\
  
    \bottomrule
    \end{tabular}
    }
  \label{tab:ablation}
\end{table}

\begin{figure*}[tbp]
\centering
  \includegraphics[width=1.0\linewidth]{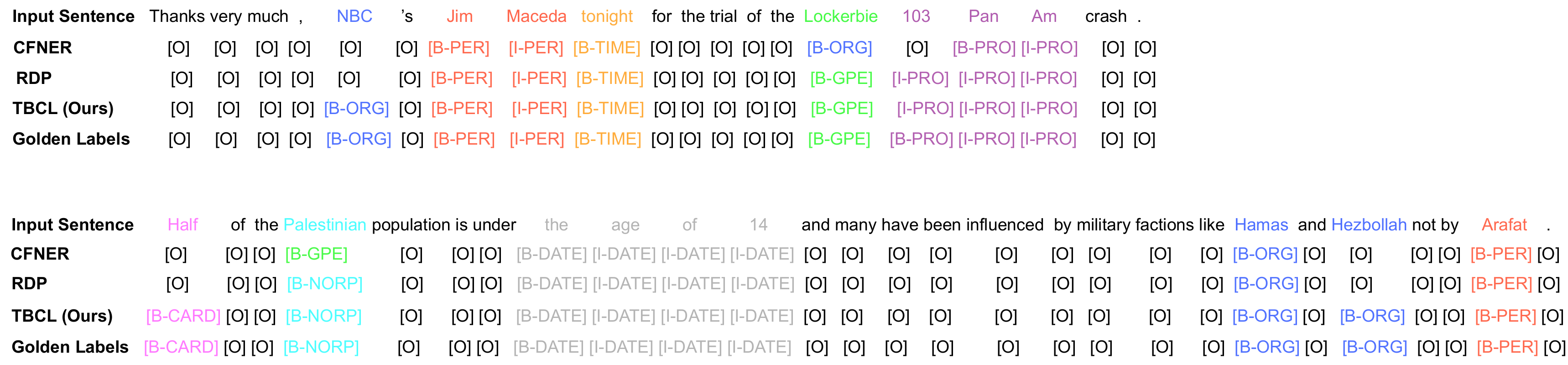}
\caption{Two real NER examples sampled from the OntoNotes5 \cite{hovy2006ontonotes} test set. 
\textbf{B-} and \textbf{I-} indicate the beginning and inside of named entities, respectively. The labels [O], \textcolor{blue}{[ORG]}, \textcolor{red}{[PER]}, \textcolor{orange}{[TIME]}, \textcolor{green}{[GPE]}, \textcolor{purple}{[PRO]}, \textcolor{magenta}{[CARD]}, \textcolor{cyan}{[NORP]}, and \textcolor{gray}{[DATE]} correspond to non-entity, \textcolor{blue}{\texttt{Organization}}, \textcolor{red}{\texttt{Person}}, \textcolor{orange}{\texttt{Time}}, \textcolor{green}{\texttt{Country/City/State}}, \textcolor{purple}{\texttt{Product}}, \textcolor{magenta}{\texttt{Cardinal}}, \textcolor{cyan}{\texttt{Nationality/Religion/Political group}}, and \textcolor{gray}{\texttt{Date}}, respectively. 
All predictions are from the last task of the FG-8-PG-2 setting. These visualizations qualitatively demonstrate the effectiveness and superiority of our proposed TBCL method.}
\label{fig:case}
\end{figure*}

\subsubsection{Effectiveness of the sentence-duplet learning scheme}

To demonstrate the effectiveness of the sentence-duplet learning scheme, we compare the performance of the Baseline model with our enhanced approach, which includes the original sentence alongside its corresponding new-entity-type-erased version (referred to as Baseline+duplet). The results, presented in Table \ref{tab:ablation}, clearly show that Baseline+duplet outperforms the Baseline, showcasing that the sentence-duplet learning scheme can mitigate the forgetting of old entity types while preventing the overfitting of new ones by correcting biased contexts present in the Baseline.
Additionally, we compare Baseline+duplet with Baseline+double to evaluate the effect of increased sample size. Baseline+duplet uses original sentences and their corresponding erased versions in each mini-batch, while Baseline+double duplicates original sentences. As shown in Table \ref{tab:ablation}, Baseline+duplet delivers superior performance compared to Baseline+double, suggesting that merely increasing the number of samples does not guarantee performance enhancement.

\subsubsection{Effect of contextual consistency loss}

We evaluate the performance of Baseline+duplet enhanced with our contextual consistency loss (denoted as Baseline+duplet+ctxc). As shown in Table \ref{tab:ablation}, the contextual consistency loss further enhances INER performance by ensuring consistency across contexts.

\subsection{Case Study (RQ3)}

{In this subsection, we present the qualitative performance of our TBCL method compared to leading baselines, including CFNER \cite{zheng2022distilling} and RDP \cite{zhang2023task}. 
We provide visual examples from the last task of the FG-8-PG-2 setting on the OntoNotes5 dataset \cite{hovy2006ontonotes}, as shown in Figure \ref{fig:case}. 
These examples demonstrate TBCL's superior ability to sequentially learn and preserve multiple entity types, highlighting its effectiveness in the incremental learning framework.}

\subsection{More Explorations (RQ4)}

\begin{figure*}[t]
\centering
  \includegraphics[width=1.0\linewidth]{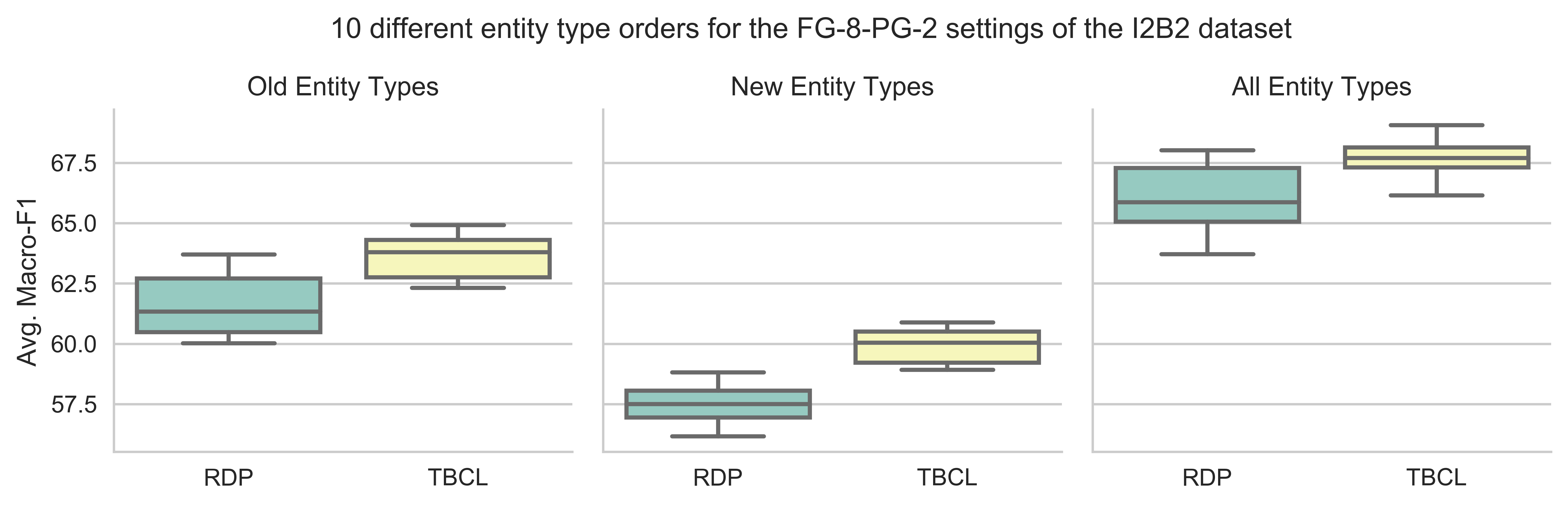} 
    \caption{The boxplots display the average Macro-F1 scores for the old, new, and all entity types across INER steps for $10$ random entity type orders. TBCL is significantly better and more stable than RDP.}
\label{fig:order}
\end{figure*}

\subsubsection{Stability Concerning Entity Type Orders}

Recent studies have shown that existing incremental learning methods may be prone to instability, with entity type ordering having a significant impact on performance \cite{DBLP:conf/cvpr/DouillardCDC21,kim2019incremental}. However, in real-world scenarios, the optimal entity type order is never known beforehand. Therefore, the performance of an ideal INER method should be as invariant to entity type order as possible. In all previous experiments, this entity type order has been kept constant (\emph{i.e.}, alphabetical order), as defined in \cite{zheng2022distilling,zhang2023task}. Here, we report results in Figure \ref{fig:order} in the form of boxplots, obtained by applying $10$ random permutations of the entity type order under the FG-8-PG-2 setting of the I2B2 dataset \cite{murphy2010serving}. Figure \ref{fig:order} (from left to right) shows the average Macro-F1 scores for the old, new, and all entity types across all INER steps. 
As shown in Figure \ref{fig:order}, the boxplot for TBCL has higher values, a more concentrated data distribution, and a smaller range of variation compared to RDP. Therefore, our proposed TBCL method performs significantly better and is more stable than the previous SOTA RDP approach.

\begin{table}[tbp]
  \centering
  \caption{{Performance on the I2B2 dataset \cite{murphy2010serving} under the FG-8-PG-2 setting with different encoder backbones. \textbf{Bold} denotes the best results for each method.}}
  \resizebox{1.0\linewidth}{!}{
    \begin{tabular}{lcccc}
    \toprule
    \multirow{2}{*}{Encoder Backbone} & \multicolumn{2}{c}{RDP} & \multicolumn{2}{c}{TBCL} \\
\cmidrule(lr){2-3} \cmidrule(lr){4-5}
    & Avg. Micro-F1 & Avg. Macro-F1 & Avg. Micro-F1 & Avg. Macro-F1 \\
    \midrule
    bert-base-cased & 80.08±0.40 & 63.72±0.71 & 82.37±1.19 & 66.51±1.32\\
    bert-large-cased & 81.25±1.30 &  65.10±1.45 & 83.39±0.82 & 68.57±1.13\\    
    roberta-large &\textbf{82.50±0.95} & \textbf{66.80±1.21} & \textbf{84.20±1.06} & \textbf{69.18±1.17} \\
    \bottomrule
    \end{tabular}
  }
  \label{tab:encoder_rdp_tbcl}
\end{table}

\begin{table*}[tbp]
\centering
\caption{Comparisons with baselines on the FEW-NERD \cite{ding2021few} dataset at each incremental step (total 11 steps). The \textcolor{deepred}{\textbf{red}} denotes the highest result, while the \textcolor{blue}{\textbf{blue}} denotes the second highest result.}
\resizebox{1.0\linewidth}{!}{%
\begin{tabular}{lccccccccccc|c}
\hline
Method & $S_1$ & $S_2$ & $S_3$ & $S_4$ & $S_5$ & $S_6$ & $S_7$ & $S_8$ & $S_9$ & $S_{10}$ & $S_{11}$ & Avg. \\
\hline

PODNet \cite{DBLP:conf/eccv/DouillardCORV20} & 48.12 & 12.84 & 9.56 & 10.21 & 6.91 & 5.72 & 4.28 & 5.01 & 4.83& 3.45& 4.12 & 10.46\\

LUCIR \cite{DBLP:conf/cvpr/HouPLWL19} & 48.12 & 45.16 & 42.87& 38.90 & 26.85 & 21.43 & 25.33& 27.99& 22.48& 20.05& 19.11 & 30.75 \\

Self-Training \cite{DBLP:conf/wacv/RosenbergHS05} & 48.12 & 40.28& 36.71& 33.92& 28.65& 25.34& 31.12& 33.85& 33.01& 31.07& 29.56& 33.78 \\

ExtendNER \cite{monaikul2021continual} & 48.12 &39.88 & 39.12& 31.05 & 28.46& 26.92& 31.77& 34.56& 32.88& 33.21& 30.85 & 34.26 \\

CFNER \cite{zheng2022distilling} & 48.12 &49.24& \textcolor{blue}{\textbf{47.65}}& \textcolor{blue}{\textbf{46.78}} &38.95& 34.87&\textcolor{blue}{\textbf{37.92}}& 40.88& \textcolor{blue}{\textbf{39.55}} & \textcolor{blue}{\textbf{41.23}}& 38.49 & 42.15 \\

RDP \cite{zhang2023task} & 48.12 & \textcolor{blue}{\textbf{50.21}} & \textcolor{deepred}{\textbf{47.83}} & 45.56& \textcolor{blue}{\textbf{41.27}} & \textcolor{blue}{\textbf{38.92}}& 37.45& \textcolor{blue}{\textbf{40.89}} & 39.12& 39.75& \textcolor{blue}{\textbf{39.04}}& \textcolor{blue}{\textbf{42.56}}\\

\hline

\textbf{TBCL (Ours)} & 48.12 & \textcolor{deepred}{\textbf{50.41}} & 47.27& \textcolor{deepred}{\textbf{47.05}} & \textcolor{deepred}{\textbf{46.12}}& \textcolor{deepred}{\textbf{43.11}} & \textcolor{deepred}{\textbf{43.95}} & \textcolor{deepred}{\textbf{43.23}} &\textcolor{deepred}{\textbf{42.11}}& \textcolor{deepred}{\textbf{41.38}} & \textcolor{deepred}{\textbf{40.71}} & \textcolor{deepred}{\textbf{44.86}} \\

\bottomrule
\end{tabular}
}
\label{tab:fewnerd_results}
\end{table*}

\subsubsection{{Larger Encoder Backbones}}

{In previous experiments, our encoder backbone was primarily based on \texttt{bert-base-cased} \cite{kenton2019bert}. 
Here, we evaluate the performance of our TBCL method alongside the previous SOTA RDP \cite{zhang2023task} using larger pre-trained language models, specifically \texttt{bert-large-cased} \cite{kenton2019bert} and \texttt{roberta-large} \cite{liu2019roberta}.
The results on the I2B2 dataset \cite{murphy2010serving} under the FG-8-PG-2 setting are presented in Table \ref{tab:encoder_rdp_tbcl}. 
As shown, both RDP and TBCL benefit from larger encoder backbones: for example, RDP's Micro-F1 improves from 80.08 to 82.50, and TBCL's from 82.37 to 84.20. 
Across all backbones, TBCL consistently outperforms RDP in both Micro-F1 and Macro-F1, demonstrating the robustness and general applicability of our method with different pre-trained encoders. 
These results confirm that TBCL not only leverages larger language models effectively but also maintains its performance advantage over strong baselines.}

\subsubsection{{Finer-grained Entity Types}}

{To further validate the effectiveness of our proposed TBCL method, we conduct experiments on the FEW-NERD dataset \cite{ding2021few}, which contains 66 fine-grained entity types. 
We adopt an FG-6-PG-6 INER setting, where the first step learns 6 entity types and each subsequent step introduces 6 new entity types, for a total of 11 incremental steps.}

{We evaluate several strong baseline methods, including {PODNet} \cite{DBLP:conf/eccv/DouillardCORV20}, {LUCIR} \cite{DBLP:conf/cvpr/HouPLWL19}, {Self-Training} \cite{DBLP:conf/wacv/RosenbergHS05}, {ExtendNER} \cite{monaikul2021continual}, {CFNER} \cite{zheng2022distilling}, and {RDP} \cite{zhang2023task}, using their publicly available implementations under the same FG-6-PG-6 setting.}

{The results are summarized in Table \ref{tab:fewnerd_results}, which reports the step-wise Macro-F1 scores as well as the overall average Macro-F1. 
As shown, our TBCL consistently outperforms all baselines across all incremental steps, demonstrating its strong capability in handling fine-grained entity types and limited-data scenarios in INER. 
Notably, TBCL achieves the highest Macro-F1 at almost all incremental steps and maintains a clear advantage in average Macro-F1 over all compared methods.}

\section{Conclusion}

In this paper, we address the biased context issue common in pseudo-labeling-based INER approach \cite{zhang2023task} by proposing a novel TBCL method. This method integrates a sentence-duplet learning scheme and a contextual consistency loss to rectify biased context correlations between tokens of old and new entity types, reducing the problems of forgetting old entity types and overfitting new ones.
We perform extensive experiments on ten INER settings across three highly recognized datasets: CoNLL2003 \cite{sang1837introduction}, I2B2 \cite{murphy2010serving}, and OntoNotes5 \cite{hovy2006ontonotes}. The results highlight the superiority of the proposed TBCL method, which surpasses previous SOTA INER approaches.

\bibliographystyle{IEEEtran}
\bibliography{reference}

@inproceedings{DBLP:conf/acl/LiYSLYCZL19,
  author    = {Xiaoya Li and
               Fan Yin and
               Zijun Sun and
               Xiayu Li and
               Arianna Yuan and
               Duo Chai and
               Mingxin Zhou and
               Jiwei Li},
  title     = {{Entity-Relation Extraction as Multi-Turn Question Answering}},
  booktitle = {Proceedings of the 57th Conference of the Association for Computational
               Linguistics, {ACL} 2019, Florence, Italy, July 28- August 2, 2019,
               Volume 1: Long Papers},
  pages     = {1340--1350},
  year      = {2019}
}

@inproceedings{DBLP:conf/emnlp/LongprePCRD021,
  author    = {Shayne Longpre and
               Kartik Perisetla and
               Anthony Chen and
               Nikhil Ramesh and
               Chris DuBois and
               Sameer Singh},
  title     = {{Entity-Based Knowledge Conflicts in Question Answering}},
  booktitle = {Proceedings of the 2021 Conference on Empirical Methods in Natural
               Language Processing, {EMNLP} 2021, Virtual Event / Punta Cana, Dominican
               Republic, 7-11 November, 2021},
  pages     = {7052--7063},
  year      = {2021}
}

@inproceedings{DBLP:conf/sigir/FetahuFRM21,
  author    = {Besnik Fetahu and
               Anjie Fang and
               Oleg Rokhlenko and
               Shervin Malmasi},
  title     = {{Gazetteer Enhanced Named Entity Recognition for Code-Mixed Web Queries}},
  booktitle = {{SIGIR} '21: The 44th International {ACM} {SIGIR} Conference on Research
               and Development in Information Retrieval, Virtual Event, Canada, July
               11-15, 2021},
  pages     = {1677--1681},
  year      = {2021}
}

@inproceedings{DBLP:conf/sigir/GuoXCL09,
  author    = {Jiafeng Guo and
               Gu Xu and
               Xueqi Cheng and
               Hang Li},
  title     = {{Named entity recognition in query}},
  booktitle = {Proceedings of the 32nd Annual International {ACM} {SIGIR} Conference
               on Research and Development in Information Retrieval, {SIGIR} 2009,
               Boston, MA, USA, July 19-23, 2009},
  pages     = {267--274},
  year      = {2009}
}

@inproceedings{DBLP:conf/aaai/MokhtariMY019,
  author    = {Shekoofeh Mokhtari and
               Ahmad Mahmoody and
               Dragomir Yankov and
               Ning Xie},
  title     = {{Tagging Address Queries in Maps Search}},
  booktitle = {The Thirty-Third {AAAI} Conference on Artificial Intelligence, {AAAI}
               2019, The Thirty-First Innovative Applications of Artificial Intelligence
               Conference, {IAAI} 2019, The Ninth {AAAI} Symposium on Educational
               Advances in Artificial Intelligence, {EAAI} 2019, Honolulu, Hawaii,
               USA, January 27 - February 1, 2019},
  pages     = {9547--9551},
  year      = {2019}
}

@inproceedings{DBLP:conf/kdd/ZhangJD0YCTHWHC21,
  author    = {Ningyu Zhang and
               Qianghuai Jia and
               Shumin Deng and
               Xiang Chen and
               Hongbin Ye and
               Hui Chen and
               Huaixiao Tou and
               Gang Huang and
               Zhao Wang and
               Nengwei Hua and
               Huajun Chen},
  title     = {{AliCG: Fine-grained and Evolvable Conceptual Graph Construction for
               Semantic Search at Alibaba}},
  booktitle = {{KDD} '21: The 27th {ACM} {SIGKDD} Conference on Knowledge Discovery
               and Data Mining, Virtual Event, Singapore, August 14-18, 2021},
  pages     = {3895--3905},
  year      = {2021}
}

@inproceedings{ma2016end,
  title={{End-to-end Sequence Labeling via Bi-directional LSTM-CNNs-CRF}},
  author={Ma, Xuezhe and Hovy, Eduard},
  booktitle={Proceedings of the Annual Meeting of the Association for Computational Linguistics},
  pages={1064--1074},
  year={2016}
}

@inproceedings{DBLP:conf/naacl/LampleBSKD16,
  author    = {Guillaume Lample and
               Miguel Ballesteros and
               Sandeep Subramanian and
               Kazuya Kawakami and
               Chris Dyer},
  title     = {{Neural Architectures for Named Entity Recognition}},
  booktitle = {{NAACL} {HLT} 2016, The 2016 Conference of the North American Chapter
               of the Association for Computational Linguistics: Human Language Technologies,
               San Diego California, USA, June 12-17, 2016},
  pages     = {260--270},
  year      = {2016},
  url       = {https://doi.org/10.18653/v1/n16-1030}
}

@article{li2020survey,
  title={{A survey on deep learning for named entity recognition}},
  author={Li, Jing and Sun, Aixin and Han, Jianglei and Li, Chenliang},
  journal={IEEE Transactions on Knowledge and Data Engineering},
  volume={34},
  number={1},
  pages={50--70},
  year={2020},
  publisher={IEEE}
}

@inproceedings{DBLP:conf/cvpr/DouillardCDC21,
  author    = {Arthur Douillard and
               Yifu Chen and
               Arnaud Dapogny and
               Matthieu Cord},
  title     = {{{PLOP:} Learning Without Forgetting for Continual Semantic Segmentation}},
  booktitle = {{IEEE} Conference on Computer Vision and Pattern Recognition, {CVPR}
               2021, virtual, June 19-25, 2021},
  pages     = {4040--4050},
  year      = {2021}
}

@inproceedings{neural_for_nlp,
  title={Neural architectures for named entity recognition},
  author={Lample, Guillaume and Ballesteros, Miguel and Subramanian, Sandeep and Kawakami, Kazuya and Dyer, Chris},
  booktitle={The North American Chapter of the Association for Computational Linguistics},
  year={2016}
}

@incollection{lifelong_learning,
  title={Lifelong learning algorithms},
  author={Thrun, Sebastian},
  booktitle={Learning to learn},
  pages={181--209},
  year={1998},
  publisher={Springer}
}

@article{continual_lifelong,
  title={Continual lifelong learning with neural networks: A review},
  author={Parisi, German I and Kemker, Ronald and Part, Jose L and Kanan, Christopher and Wermter, Stefan},
  journal={Neural Networks},
  volume={113},
  pages={54--71},
  year={2019},
  publisher={Elsevier}
}

@article{catastrophic_1,
  title={{Catastrophic interference in connectionist networks: The sequential learning problem}},
  author={McCloskey, Michael and Cohen, Neal J},
  journal={Psychology of learning and motivation},
  volume={24},
  pages={109--165},
  year={1989},
  publisher={Elsevier}
}

@article{catastrophic_2,
  title={{Catastrophic forgetting, rehearsal and pseudorehearsal}},
  author={Robins, Anthony},
  journal={Connection Science},
  volume={7},
  number={2},
  pages={123--146},
  year={1995},
  publisher={Taylor \& Francis}
}

@article{catastrophic_3,
  title={{An empirical investigation of catastrophic forgetting in gradient-based neural networks}},
  author={Goodfellow, Ian J and Mirza, Mehdi and Xiao, Da and Courville, Aaron and Bengio, Yoshua},
  journal={arXiv preprint arXiv:1312.6211},
  year={2013}
}

@article{catastrophic_4,
  title={{Overcoming catastrophic forgetting in neural networks}},
  author={Kirkpatrick, James and Pascanu, Razvan and Rabinowitz, Neil and Veness, Joel and Desjardins, Guillaume and Rusu, Andrei A and Milan, Kieran and Quan, John and Ramalho, Tiago and Grabska-Barwinska, Agnieszka and others},
  journal={Proceedings of the national academy of sciences},
  volume={114},
  number={13},
  pages={3521--3526},
  year={2017},
  publisher={National Acad Sciences}
}

@inproceedings{zheng2022distilling,
  title={{Distilling Causal Effect from Miscellaneous Other-Class for Continual Named Entity Recognition}},
  author={Zheng, Junhao and Liang, Zhanxian and Chen, Haibin and Ma, Qianli},
  booktitle={Proceedings of the Conference on Empirical Methods in Natural Language Processing},
  year={2022}
}

@article{murphy2010serving,
  title={{Serving the enterprise and beyond with informatics for integrating biology and the bedside (i2b2)}},
  author={Murphy, Shawn N and Weber, Griffin and Mendis, Michael and Gainer, Vivian and Chueh, Henry C and Churchill, Susanne and Kohane, Isaac},
  journal={Journal of the American Medical Informatics Association},
  volume={17},
  number={2},
  pages={124--130},
  year={2010},
  publisher={BMJ Group}
}

@inproceedings{hovy2006ontonotes,
  title={{OntoNotes: the 90\% solution}},
  author={Hovy, Eduard and Marcus, Mitch and Palmer, Martha and Ramshaw, Lance and Weischedel, Ralph},
  booktitle={Proceedings of the human language technology conference of the NAACL, Companion Volume: Short Papers},
  pages={57--60},
  year={2006}
}

@article{sang1837introduction,
  title={{Introduction to the CoNLL-2003 Shared Task: Language-Independent Named Entity Recognition}},
  author={Sang, Erik F Tjong Kim and De Meulder, Fien},
  journal={Development},
  volume={922},
  pages={1341},
  year={1837}
}

@inproceedings{monaikul2021continual,
  title={{Continual learning for named entity recognition}},
  author={Monaikul, Natawut and Castellucci, Giuseppe and Filice, Simone and Rokhlenko, Oleg},
  booktitle={Proceedings of the AAAI Conference on Artificial Intelligence},
  volume={35},
  number={15},
  pages={13570--13577},
  year={2021}
}

@inproceedings{xia2022learn,
  title={{Learn and Review: Enhancing Continual Named Entity Recognition via Reviewing Synthetic Samples}},
  author={Xia, Yu and Wang, Quan and Lyu, Yajuan and Zhu, Yong and Wu, Wenhao and Li, Sujian and Dai, Dai},
  booktitle={Findings of the Association for Computational Linguistics: ACL 2022},
  pages={2291--2300},
  year={2022}
}

@inproceedings{kenton2019bert,
  title={{BERT: Pre-training of Deep Bidirectional Transformers for Language Understanding}},
  author={Kenton, Jacob Devlin Ming-Wei Chang and Toutanova, Lee Kristina},
  booktitle={Proceedings of NAACL-HLT},
  pages={4171--4186},
  year={2019}
}

@article{paszke2019pytorch,
  title={{Pytorch: An imperative style, high-performance deep learning library}},
  author={Paszke, Adam and Gross, Sam and Massa, Francisco and Lerer, Adam and Bradbury, James and Chanan, Gregory and Killeen, Trevor and Lin, Zeming and Gimelshein, Natalia and Antiga, Luca and others},
  journal={Advances in neural information processing systems},
  volume={32},
  year={2019}
}

@article{wolf2019huggingface,
  title={{Huggingface's transformers: State-of-the-art natural language processing}},
  author={Wolf, Thomas and Debut, Lysandre and Sanh, Victor and Chaumond, Julien and Delangue, Clement and Moi, Anthony and Cistac, Pierric and Rault, Tim and Louf, R{\'e}mi and Funtowicz, Morgan and others},
  journal={arXiv preprint arXiv:1910.03771},
  year={2019}
}

@inproceedings{KD,
  title={{Distilling the knowledge in a neural network}},
  author={Hinton, Geoffrey and Vinyals, Oriol and Dean, Jeff and others},
  booktitle={NIPS Deep Learning and Representation Learning Workshop},
  year={2015}
}

@article{DBLP:journals/corr/abs-1909-08383,
  author    = {Matthias De Lange and
               Rahaf Aljundi and
               Marc Masana and
               Sarah Parisot and
               Xu Jia and
               Ales Leonardis and
               Gregory G. Slabaugh and
               Tinne Tuytelaars},
  title     = {{Continual learning: {A} comparative study on how to defy forgetting
               in classification tasks}},
  journal   = {CoRR},
  volume    = {abs/1909.08383},
  year      = {2019}
}

@inproceedings{DBLP:conf/cvpr/HouPLWL19,
  author    = {Saihui Hou and
               Xinyu Pan and
               Chen Change Loy and
               Zilei Wang and
               Dahua Lin},
  title     = {{Learning a Unified Classifier Incrementally via Rebalancing}},
  booktitle = {{IEEE} Conference on Computer Vision and Pattern Recognition, {CVPR}
               2019, Long Beach, CA, USA, June 16-20, 2019},
  pages     = {831--839},
  year      = {2019}
}

@inproceedings{DBLP:conf/eccv/DouillardCORV20,
  author    = {Arthur Douillard and
               Matthieu Cord and
               Charles Ollion and
               Thomas Robert and
               Eduardo Valle},
  title     = {{PODNet: Pooled Outputs Distillation for Small-Tasks Incremental Learning}},
  booktitle = {Computer Vision - {ECCV} 2020 - 16th European Conference, Glasgow,
               UK, August 23-28, 2020, Proceedings, Part {XX}},
  pages     = {86--102},
  year      = {2020}
}

@inproceedings{DBLP:conf/wacv/RosenbergHS05,
  author    = {Chuck Rosenberg and
               Martial Hebert and
               Henry Schneiderman},
  title     = {{Semi-Supervised Self-Training of Object Detection Models}},
  booktitle = {7th {IEEE} Workshop on Applications of Computer Vision / {IEEE} Workshop
               on Motion and Video Computing {(WACV/MOTION} 2005), 5-7 January 2005,
               Breckenridge, CO, {USA}},
  pages     = {29--36},
  year      = {2005}
}

@article{lifelong_machine,
  title={{Lifelong machine learning}},
  author={Chen, Zhiyuan and Liu, Bing},
  journal={Synthesis Lectures on Artificial Intelligence and Machine Learning},
  volume={12},
  number={3},
  pages={1--207},
  year={2018},
  publisher={Morgan \& Claypool Publishers}
}

@article{EWC,
  title={{Overcoming catastrophic forgetting in neural networks}},
  author={Kirkpatrick, James and Pascanu, Razvan and Rabinowitz, Neil and Veness, Joel and Desjardins, Guillaume and Rusu, Andrei A and Milan, Kieran and Quan, John and Ramalho, Tiago and Grabska-Barwinska, Agnieszka and others},
  journal={Proceedings of the national academy of sciences},
  volume={114},
  number={13},
  pages={3521--3526},
  year={2017},
  publisher={National Acad Sciences}
}

@InProceedings{SI,
author = {Aljundi, Rahaf and Babiloni, Francesca and Elhoseiny, Mohamed and Rohrbach, Marcus and Tuytelaars, Tinne},
title = {{Memory Aware Synapses: Learning what (not) to forget }},
booktitle = {Proceedings of the European Conference on Computer Vision (ECCV)},
month = {September},
year = {2018}
}

@inproceedings{MAS,
  title={{Continual learning through synaptic intelligence}},
  author={Zenke, Friedemann and Poole, Ben and Ganguli, Surya},
  booktitle={International Conference on Machine Learning (ICML)},
  pages={3987--3995},
  year={2017},
  organization={PMLR}
}

@inproceedings{ONLINE-EWC,
  title={{Progress \& compress: A scalable framework for continual learning}},
  author={Schwarz, Jonathan and Czarnecki, Wojciech and Luketina, Jelena and Grabska-Barwinska, Agnieszka and Teh, Yee Whye and Pascanu, Razvan and Hadsell, Raia},
  booktitle={International Conference on Machine Learning (ICML)},
  pages={4528--4537},
  year={2018},
  organization={PMLR}
}

@ARTICLE{LWF,
  author={Li, Zhizhong and Hoiem, Derek},
  journal={IEEE Transactions on Pattern Analysis and Machine Intelligence (TPAMI)}, 
  title={{Learning without Forgetting}}, 
  year={2018},
  volume={40},
  number={12},
  pages={2935-2947},
  doi={10.1109/TPAMI.2017.2773081}}

@InProceedings{CLASSIFIER,
author = {Hou, Saihui and Pan, Xinyu and Loy, Chen Change and Wang, Zilei and Lin, Dahua},
title = {{Learning a Unified Classifier Incrementally via Rebalancing}},
booktitle = {Proceedings of the IEEE/CVF Conference on Computer Vision and Pattern Recognition (CVPR)},
month = {June},
year = {2019}
}

@InProceedings{icarl,
author = {Rebuffi, Sylvestre-Alvise and Kolesnikov, Alexander and Sperl, Georg and Lampert, Christoph H.},
title = {{iCaRL: Incremental Classifier and Representation Learning}},
booktitle = {Proceedings of the IEEE Conference on Computer Vision and Pattern Recognition (CVPR)},
month = {July},
year = {2017}
}

@inproceedings{GEM,
 author = {Lopez-Paz, David and Ranzato, Marc\textquotesingle Aurelio},
 booktitle = {Advances in Neural Information Processing Systems},
 editor = {I. Guyon and U. Von Luxburg and S. Bengio and H. Wallach and R. Fergus and S. Vishwanathan and R. Garnett},
 pages = {},
 publisher = {Curran Associates, Inc.},
 title = {{Gradient Episodic Memory for Continual Learning}},
 url = {https://proceedings.neurips.cc/paper/2017/file/f87522788a2be2d171666752f97ddebb-Paper.pdf},
 volume = {30},
 year = {2017}
}

@inproceedings{DGR,
 author = {Shin, Hanul and Lee, Jung Kwon and Kim, Jaehong and Kim, Jiwon},
 booktitle = {Advances in Neural Information Processing Systems},
 editor = {I. Guyon and U. Von Luxburg and S. Bengio and H. Wallach and R. Fergus and S. Vishwanathan and R. Garnett},
 pages = {},
 publisher = {Curran Associates, Inc.},
 title = {{Continual Learning with Deep Generative Replay}},
 url = {https://proceedings.neurips.cc/paper/2017/file/0efbe98067c6c73dba1250d2beaa81f9-Paper.pdf},
 volume = {30},
 year = {2017}
}

@InProceedings{PACKNET,
author = {Mallya, Arun and Lazebnik, Svetlana},
title = {{PackNet: Adding Multiple Tasks to a Single Network by Iterative Pruning}},
booktitle = {Proceedings of the IEEE Conference on Computer Vision and Pattern Recognition (CVPR)},
month = {June},
year = {2018}
}

@ARTICLE{DA,
  author={Rosenfeld, Amir and Tsotsos, John K.},
  journal={IEEE Transactions on Pattern Analysis and Machine Intelligence (TPAMI)}, 
  title={{Incremental Learning Through Deep Adaptation}}, 
  year={2020},
  volume={42},
  number={3},
  pages={651-663},
  doi={10.1109/TPAMI.2018.2884462}}

@inproceedings{DEN,
  title={Lifelong Learning with Dynamically Expandable Networks},
  author={Yoon, Jaehong and Yang, Eunho and Lee, Jeongtae and Hwang, Sung Ju},
  year={2018},
  booktitle={Proceedings of the International Conference on Learning Representations (ICLR)}
}

@InProceedings{dong2023federated,
    author = {Dong, Jiahua and Zhang, Duzhen and Cong, Yang and Cong, Wei and Ding, Henghui and Dai, Dengxin},
    title = {{Federated Incremental Semantic Segmentation}},
    booktitle = {IEEE/CVF Conference on Computer Vision and Pattern Recognition (CVPR)},
    month = {June},
    year = {2023},
}

@inproceedings{zhang2023decomposing,
  title={{Decomposing Logits Distillation for Incremental Named Entity Recognition}},
  author={Zhang, Duzhen and Yu, Yahan and Chen, Feilong and Chen, Xiuyi},
  booktitle={Proceedings of the 46th International ACM SIGIR Conference on Research and Development in Information Retrieval},
  pages={1919--1923},
  year={2023}
}

@inproceedings{zhang2023continual,
  title={{Continual Named Entity Recognition without Catastrophic Forgetting}},
  author={Zhang, Duzhen and Cong, Wei and Dong, Jiahua and Yu, Yahan and Chen, Xiuyi and Zhang, Yonggang and Fang, Zhen},
  booktitle={Proceedings of the 2023 Conference on Empirical Methods in Natural Language Processing},
  pages={8186--8197},
  year={2023}
}

@inproceedings{zhang2023task,
  title={{Task relation distillation and prototypical pseudo label for incremental named entity recognition}},
  author={Zhang, Duzhen and Li, Hongliu and Cong, Wei and Xu, Rongtao and Dong, Jiahua and Chen, Xiuyi},
  booktitle={Proceedings of the 32nd ACM International Conference on Information and Knowledge Management (CIKM)},
  pages={3319--3329},
  year={2023}
}

@article{hinton2015distilling,
  title={{Distilling the knowledge in a neural network}},
  author={Hinton, Geoffrey and Vinyals, Oriol and Dean, Jeff and others},
  journal={arXiv preprint arXiv:1503.02531},
  volume={2},
  number={7},
  year={2015}
}

@inproceedings{ma2023learning,
  title={{Learning “O” Helps for Learning More: Handling the Unlabeled Entity Problem for Class-incremental NER}},
  author={Ma, Ruotian and Chen, Xuanting and Lin, Zhang and Zhou, Xin and Wang, Junzhe and Gui, Tao and Zhang, Qi and Gao, Xiang and Chen, Yun Wen},
  booktitle={Proceedings of the 61st Annual Meeting of the Association for Computational Linguistics (Volume 1: Long Papers)},
  pages={5959--5979},
  year={2023}
}

@inproceedings{chen2023skd,
  title={{SKD-NER: Continual Named Entity Recognition via Span-based Knowledge Distillation with Reinforcement Learning}},
  author={Chen, Yi and He, Liang},
  booktitle={Proceedings of the 2023 Conference on Empirical Methods in Natural Language Processing},
  pages={6689--6700},
  year={2023}
}

@inproceedings{zhao2022rbc,
  title={{Rbc: Rectifying the biased context in continual semantic segmentation}},
  author={Zhao, Hanbin and Yang, Fengyu and Fu, Xinghe and Li, Xi},
  booktitle={European Conference on Computer Vision},
  pages={55--72},
  year={2022},
  organization={Springer}
}

@article{kim2019incremental,
  title={{Incremental learning with maximum entropy regularization: Rethinking forgetting and intransigence}},
  author={Kim, Dahyun and Bae, Jihwan and Jo, Yeonsik and Choi, Jonghyun},
  journal={arXiv preprint arXiv:1902.00829},
  year={2019}
}

@article{liu2019roberta,
  title={{Roberta: A robustly optimized bert pretraining approach}},
  author={Liu, Yinhan and Ott, Myle and Goyal, Naman and Du, Jingfei and Joshi, Mandar and Chen, Danqi and Levy, Omer and Lewis, Mike and Zettlemoyer, Luke and Stoyanov, Veselin},
  journal={arXiv preprint arXiv:1907.11692},
  year={2019}
}

@inproceedings{ding2021few,
  title={{Few-nerd: A few-shot named entity recognition dataset}},
  author={Ding, Ning and Xu, Guangwei and Chen, Yulin and Wang, Xiaobin and Han, Xu and Xie, Pengjun and Zheng, Hai-Tao and Liu, Zhiyuan},
  booktitle={Proceedings of the 59th Annual Meeting of the Association for Computational Linguistics and the 11th International Joint Conference on Natural Language Processing (Volume 1: Long Papers)},
  pages={3198--3213},
  year={2021}
}

@article{zhang2025exploring,
  title={{Exploring Stability-Plasticity Trade-offs for Continual Named Entity Recognition}},
  author={Zhang, Duzhen and Li, Chenxing and Dong, Jiahua and Liu, Qi and Yu, Dong},
  journal={IEEE Transactions on Audio, Speech and Language Processing},
  year={2025},
  publisher={IEEE}
}

@inproceedings{yu2024flexible,
  title={{Flexible weight tuning and weight fusion strategies for continual named entity recognition}},
  author={Yu, Yahan and Zhang, Duzhen and Chen, Xiuyi and Chu, Chenhui},
  booktitle={Findings of the Association for Computational Linguistics: ACL 2024},
  pages={1351--1358},
  year={2024}
}

\begin{IEEEbiography}[{\includegraphics[width=1in,height=1.25in,clip,keepaspectratio]{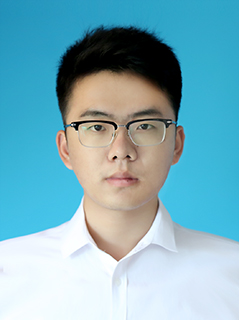}}]{Duzhen Zhang} received his B.Sc. degree from Shandong University in June 2019. He completed his Ph.D. at the Institute of Automation, Chinese Academy of Sciences in June 2024. Since September 2024, he has been a postdoctoral researcher at the Mohamed bin Zayed University of Artificial Intelligence. His current research interests include large language models, continual learning, multi-modal learning, and AI for science.
\end{IEEEbiography} 

\begin{IEEEbiography}
[{\includegraphics[width=1in,height=1.25in,clip,keepaspectratio]{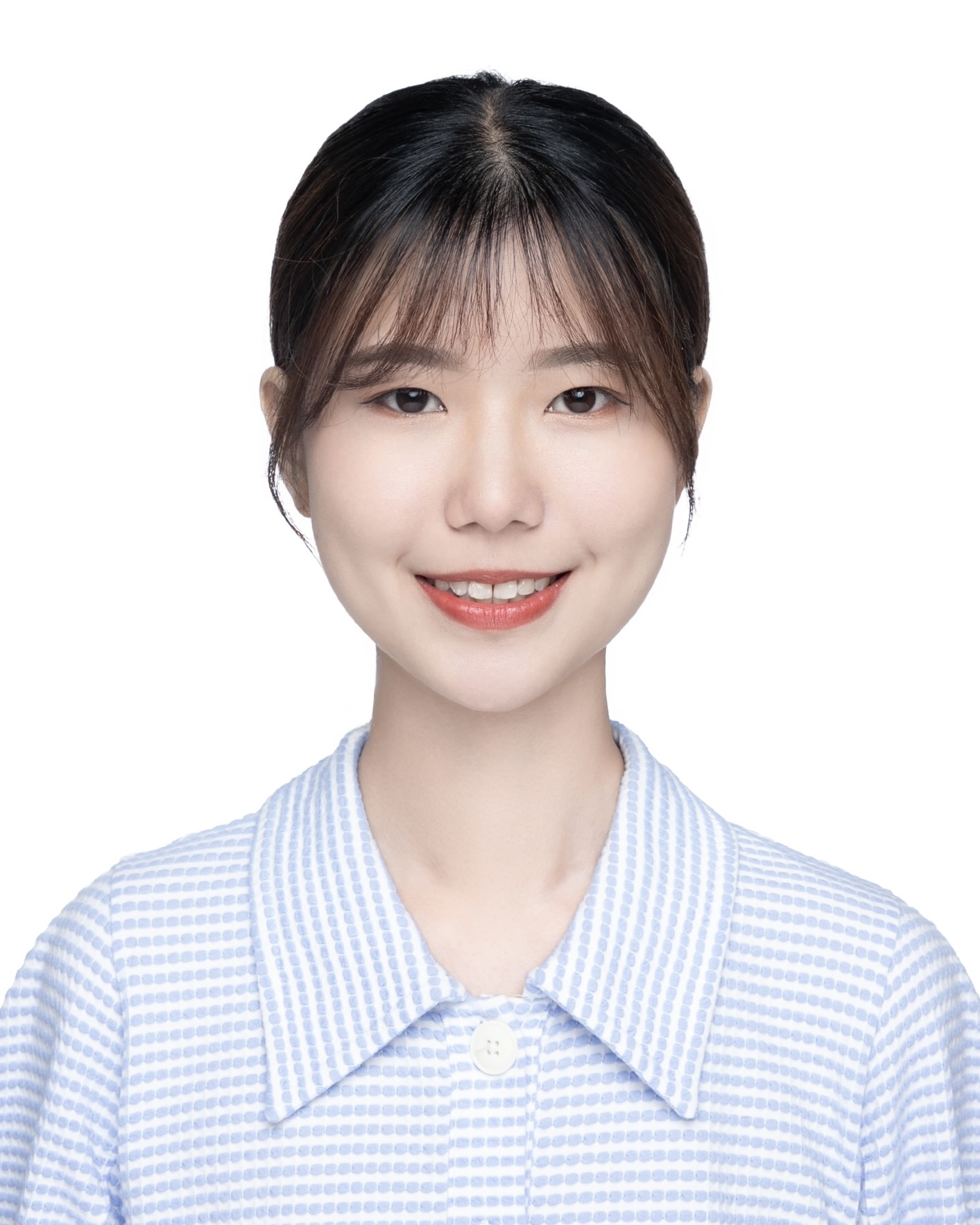}}]{Yahan Yu} received her B.S. degree from Nanjing University of Aeronautics and Astronautics in 2020 and her M.S. degree in Control Science from University of Chinese Academy of Sciences in 2023.
She is currently a Ph.D. student at Kyoto University. Her research interests include natural language processing and continual learning.
\end{IEEEbiography}

\begin{IEEEbiography}[{\includegraphics[width=1in,height=1.2in,clip]{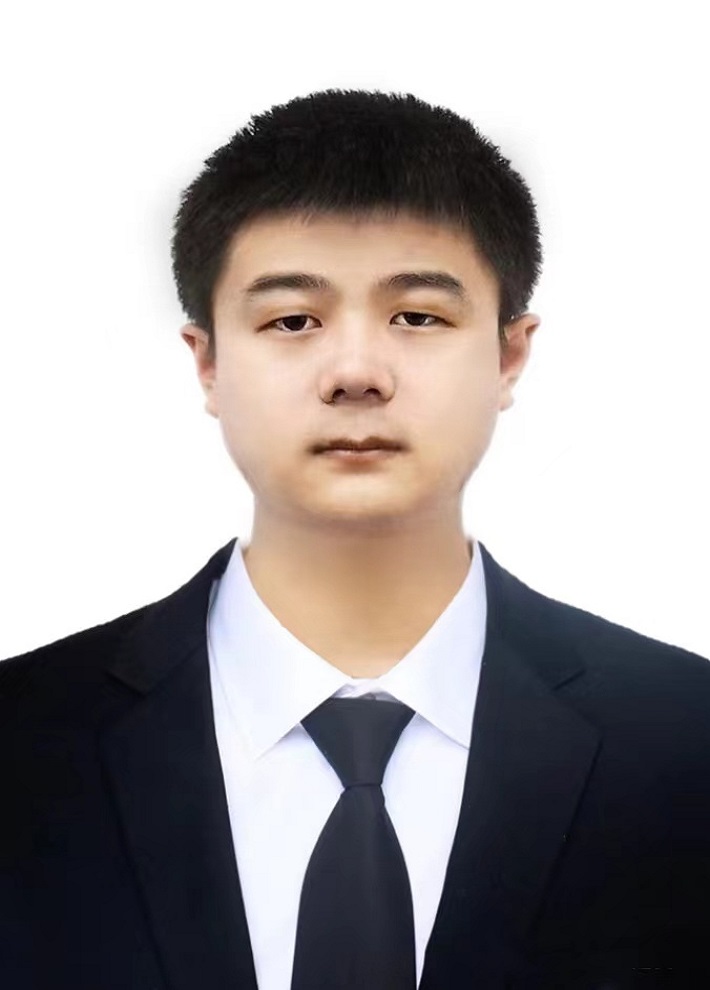}}]{Xiuyi Chen} received his Ph.D. degree (2022) in Pattern Recognition and Intelligent System from Institute of Automation, Chinese Academy of Sciences, advised by Prof. Bo Xu. Previously, he received the B.Sc. degree (2017) from JiLin University. His current interests include Cross-modal Retrieval, Multimodal Learning, Dialogue System, and Knowledge-Grounded Generation.
\end{IEEEbiography} 

\begin{IEEEbiography}[{\includegraphics[width=1in,height=1.25in,clip,keepaspectratio]{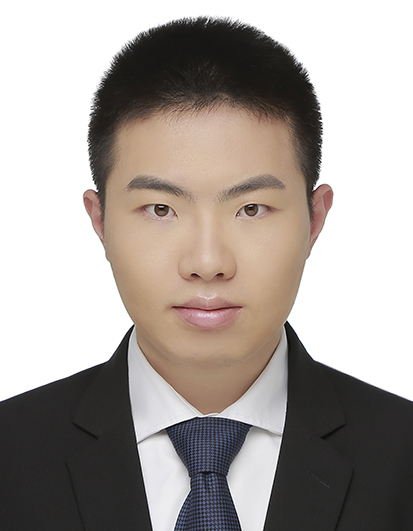}}]{Chenxing Li} received his B.Sc. degree at the North China Electric Power University, China, in 2015. He completed his Ph.D. at the Institute of Automation, Chinese Academy of Sciences in 2020. His current research interests include far-field speech recognition, speech enhancement, and speech separation.
\end{IEEEbiography}

\begin{IEEEbiography}[{\includegraphics[width=1in,height=1.25in,clip,keepaspectratio]{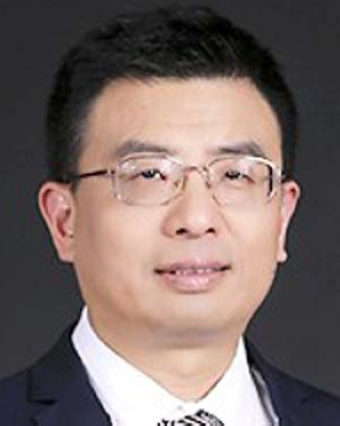}}]{Dong Yu} (Fellow, IEEE) with the Tencent AI Lab as a distinguished scientist and vice general Manager. Prior to joining Tencent in 2017, he was a Principal Researcher with Microsoft Research (Redmond), where he has been since 1998. He has authored or coauthored two monographs and more than 300 papers. His research interests include speech recognition and processing and natural language processing. His works have been widely cited and recognized by the prestigious IEEE Signal Processing Society best transaction paper award in 2013, 2016, 2020, and 2022, the 2021 NAACL best long paper award, 2022 IEEE Signal Processing Magazine best paper award, and 2022 IEEE Signal Processing Magazine best column award. Dr. Dong Yu was the Chair of the IEEE Speech and Language Processing Technical Committee during 2021–2022. He was on the editorial boards of numerous journals and magazines, as well as on the organizing and technical committees of various conferences and workshops. He is currently an ACM/IEEE/ISCA Fellow.
\end{IEEEbiography}

\end{document}